\documentclass[sigconf]{acmart}
\AtBeginDocument{%
  \providecommand\BibTeX{{%
    \normalfont B\kern-0.5em{\scshape i\kern-0.25em b}\kern-0.8em\TeX}}}

\copyrightyear{2024}
\acmYear{2024}
\setcopyright{rightsretained}
\acmConference[KDD '24]{Proceedings of the 30th ACM SIGKDD Conference on Knowledge Discovery and Data Mining}{August 25--29, 2024}{Barcelona, Spain}
\acmBooktitle{Proceedings of the 30th ACM SIGKDD Conference on Knowledge Discovery and Data Mining (KDD '24), August 25--29, 2024, Barcelona, Spain}
\acmDOI{10.1145/3637528.3671993}
\acmISBN{979-8-4007-0490-1/24/08}
\usepackage[ruled,vlined,linesnumbered]{algorithm2e}
\usepackage{amsmath}
\usepackage{booktabs}
\usepackage{multirow}
\usepackage{subcaption}
\newcommand{\method}{RC-Mixup}
\newcommand{\cmixup}{C-Mixup}
\usepackage{amsmath,amsfonts,bm}

\def\eqref#1{equation~\ref{#1}}
\def\1{\bm{1}}
\newcommand{\train}{\mathcal{D}}

\DeclareMathAlphabet{\mathsfit}{\encodingdefault}{\sfdefault}{m}{sl}
\SetMathAlphabet{\mathsfit}{bold}{\encodingdefault}{\sfdefault}{bx}{n}

\DeclareMathOperator*{\argmin}{arg\,min}

\begin{document}

\title[A Data Augmentation Strategy against Noisy Data for Regression]{\method{}: A Data Augmentation Strategy against Noisy Data \\for Regression Tasks}

\author{Seong-Hyeon Hwang}
\email{sh.hwang@kaist.ac.kr}
\affiliation{
\institution{KAIST}
\city{Daejeon}
\country{Republic of Korea}
}

\author{Minsu Kim}
\email{ms716@kaist.ac.kr}
\affiliation{
\institution{KAIST}
\city{Daejeon}
\country{Republic of Korea}
}

\author{Steven Euijong Whang}
\authornote{Corresponding author}
\email{swhang@kaist.ac.kr}
\affiliation{
\institution{KAIST}
\city{Daejeon}
\country{Republic of Korea}
}

% \renewcommand{\shortauthors}{Hwang et al.}
\renewcommand{\shortauthors}{Seong-Hyeon Hwang et al.}

\begin{abstract}
We study the problem of robust data augmentation for regression tasks in the presence of noisy data. Data augmentation is essential for generalizing deep learning models, but most of the techniques like the popular Mixup are primarily designed for classification tasks on image data. Recently, there are also Mixup techniques that are specialized to regression tasks like \cmixup{}. In comparison to Mixup, which takes linear interpolations of pairs of samples, \cmixup{} is more selective in which samples to mix based on their label distances for better regression performance. However, \cmixup{} does not distinguish noisy versus clean samples, which can be problematic when mixing and lead to suboptimal model performance. At the same time, robust training has been heavily studied where the goal is to train accurate models against noisy data through multiple rounds of model training. We thus propose our data augmentation strategy \method{}, which {\em tightly integrates \cmixup{} with multi-round robust training methods for a synergistic effect}. In particular, \cmixup{} improves robust training in identifying clean data, while robust training provides cleaner data to \cmixup{} for it to perform better. A key advantage of \method{} is that it is {\em data-centric} where the robust model training algorithm itself does not need to be modified, but can simply benefit from data mixing. We show in our experiments that \method{} significantly outperforms \cmixup{} and robust training baselines on noisy data benchmarks and can be integrated with various robust training methods.
\end{abstract}




\maketitle

\section{Introduction}
Deep learning is widely used in applications that perform regression tasks including manufacturing, climate prediction, and finance. However, one of the challenges is a lack of enough training data, and data augmentation techniques have been proposed as a solution for better generalizing the trained models. A representative technique is Mixup\,\citep{DBLP:conf/iclr/ZhangCDL18, DBLP:conf/iccv/YunHCOYC19, kim2020puzzle, kim2021comixup}, which mixes two samples by linear interpolation to estimate the label of any sample in between. However, Mixup is primarily designed for classification tasks mainly on image data along with most of the other data augmentation techniques\,\citep{DBLP:conf/nips/GoodfellowPMXWOCB14,DBLP:journals/corr/KingmaW13} and does not readily perform well on regression tasks where the goal is to predict real numbers. 

Recently, there is an increasing literature on data augmentation designed for regression tasks, and \cmixup{}\,\cite{DBLP:conf/nips/YaoWZZF22} is the state-of-the-art Mixup-based method. In order to avoid arbitrarily-incorrect labels, each sample is mixed with a neighboring sample in terms of label distance where the selection follows the proposed sampling probability calculated using a Gaussian kernel that is configured using a bandwidth parameter. The larger the parameter the wider the distribution, which means that mixing can be done with distant neighbors in terms of label distance. In a sense, the bandwidth indicates whether Mixup should be performed, and to what degree. \cmixup{} has theoretical guarantees for generalization and handling correlation shifts in the data.

At the same time, robustness against noise is becoming increasingly important for regression tasks. For example, in semiconductor manufacturing, the layer thickness in a 3D semiconductor needs to be predicted for defect detection using a model. In this case, labels (e.g., layer thickness) can be noisy due to erroneous or malfunctioning measurement equipment, leading to a decline in prediction model performance and thus revenue. Hence, global semiconductor companies make great efforts to ensure that their models are robust against noise. To address this challenge, there is a recent line of multi-round robust training~\cite{DBLP:conf/icml/ShenS19,DBLP:conf/icml/SongK019} where noisy samples are removed or fixed based on their loss values through multiple model trainings.

\begin{figure*}[t]
\centering
\includegraphics[width=0.9\textwidth]{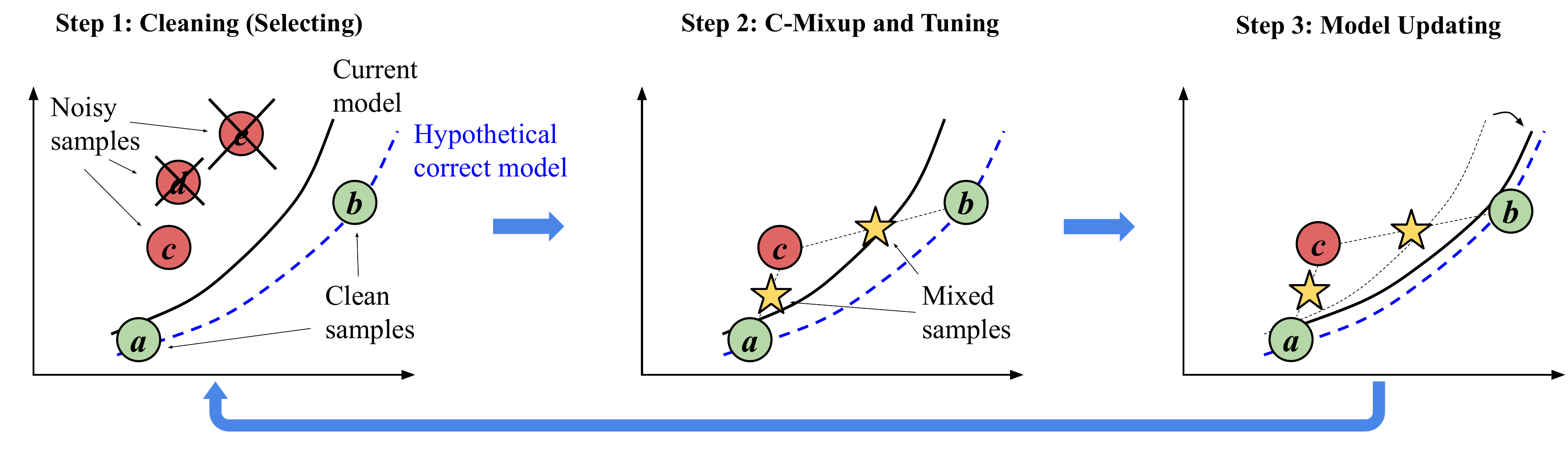}
\caption{\method{} tightly integrates \cmixup{} with multi-round robust training techniques for a synergistic effect: \cmixup{} improves robust training in identifying clean data, while robust training provides (intermediate) clean data for \cmixup{}. Suppose the x-axis is the only feature, and the y-axis is the label. Also, there are two clean samples $a$ and $b$ and three noisy samples $c$, $d$, and $e$. In Step 1, suppose that cleaning removes $d$ and $e$ (the exact outcome depends on the robust training technique). In Step 2, we perform \cmixup{} possibly with bandwidth tuning to generate mixed samples. Here we mix the sample pairs ($a$, $c$) and ($b$, $c$) to generate the mixed samples denoted as star shapes. Notice that \cmixup{} selectively mixes samples that have closer labels, so in this example ($a$, $b$) are not mixed. In Step 3, the augmented samples can be used to train an improved regression model, which can then be used for better cleaning in the next round.}
\label{fig:rcmixup}
\end{figure*}

We thus contend that integrating \cmixup{} with robust training methods is desirable, but this is not trivial. \cmixup{} is not explicitly designed to be robust against noisy data and may be prone to incorrect mixing because any out-of-distribution samples are mixed just the same way as in-distribution samples. \cmixup{} is robust against correlation shifts, which assume the same data distribution, but unfortunately cannot cope with noise in general. More fundamentally, data augmentation is to add data, while robust training is to clean (select or refurbish) data, so the two seem to even be contradictory operations. A na\"ive approach is to run the two methods in sequence, e.g., run \cmixup{} and then robust training or in the other ordering. However, either \cmixup{} will end up running on noisy data or robust training would not benefit from augmented data.

We propose the novel data augmentation strategy of {\em tightly integrating \cmixup{} and multi-round robust training for a synergistic effect} (see Figure~\ref{fig:rcmixup}). We call our framework \method{} to emphasize the robustness of \cmixup{}. Each robust training round typically consists of cleaning (Step 1) and model updating (Step 3) steps. Between these two steps, we run \cmixup{} (Step 2) so that it benefits from the intermediate clean data that is identified by the cleaning. In addition, the model updating step now benefits from the augmented data produced by \cmixup{} and produces a more accurate regression model that can clean data better. Another benefit of this integration is that it is {\em data-centric} where the robust training algorithm itself does not need to be modified because \method{} is only augmenting the data. Hence, \method{} is compatible with any existing multi-round robust training algorithm like Iterative Trimmed Loss Minimization (ITLM)\,\cite{DBLP:conf/icml/ShenS19}, O2U-Net\,\cite{DBLP:conf/iccv/HuangQJZ19}, and SELFIE\,\cite{DBLP:conf/icml/SongK019}. A technical challenge is efficiency where we would like to keep \cmixup{}'s bandwidth up-to-date during the multiple rounds in robust training. We propose to periodically update the bandwidth, but doing so reliably by evaluating candidate bandwidth values for several robust training rounds before choosing the one to use. 
We can optionally speedup this process by simply updating the bandwidth in one direction with some tradeoff in performance as well.

We perform extensive experiments of \method{} on various regression benchmarks and show how it significantly outperforms \cmixup{} in terms of robustness against noise. While \method{} utilizes a small validation set, we show it is sufficient to use the robust training to generate one from the training set. In addition, \method{} also outperforms existing robust training techniques that do not augment their data.

\textbf{Summary of Contributions:} (1) We propose \method{}, the first selective mixing framework for regression tasks that is also robust against noisy data. (2) We tightly integrate the state-of-the-art \cmixup{} with multi-round robust training techniques where \cmixup{} utilizes intermediate clean data and has a synergistic effect with robust training. (3) We perform extensive experiments and show how \method{} significantly outperforms baselines by utilizing this synergy on noisy real and synthetic datasets.

\section{Background}

\paragraph{Notations}

Let $\train = \{(x, y)\} \sim P$ be the training set where $x \in X$ is a $d$-dimensional input sample, and $y \in Y$ is an $e$-dimensional label. Let $\train^v = \{(x^v, y^v)\} \sim P^t$ be the validation set, where $P^t$ is the distribution of the test set. Let $\theta$ model parameters, $f_\theta$ be a regression model, and $\ell_\theta$ the loss function that returns a performance score comparing $f_\theta(x)$ with the true label $y$ using Mean Squared Error (MSE) loss.

\paragraph{Mixup}

Mixup\,\citep{DBLP:conf/iclr/ZhangCDL18} takes a linear interpolation between any pair of samples ${x}_i$ and ${x}_j$ with the labels ${y}_i$ and ${y}_j$ to produce the new sample $\lambda {x}_i + (1 - \lambda) {x}_j$ with the label $\lambda {y}_i + (1 - \lambda) {y}_j$ where $\lambda \sim Beta(\alpha, \alpha)$. According to \citet{DBLP:conf/iclr/ZhangCDL18}, mixing all samples outperforms Empirical Risk Minimization on many classification datasets. This strategy works well for classification where the labels are one-hot encoded where many samples may have the same label. In regression, however, such linear interpolations are not suitable as labels are in a continuous space instead of a discrete space where one-hot encodings cannot be used. Here two distant samples may have labels that are arbitrarily different, and simply mixing them could result in intermediate labels that are very different than the actual label.

\paragraph{\cmixup{}}

\cmixup{}\,\cite{DBLP:conf/nips/YaoWZZF22} overcomes the limitation of Mixup and is the state-of-the-art approach for regression tasks on clean data. \cmixup{} proposes a sampling probability distribution for each sample based on the label distance between a neighbor using a symmetric Gaussian kernel. \cmixup{} selects a sample to mix following a sampling probability and generates a new sample and label pair similar to Mixup. The sampling probability introduced by \cmixup{} is: 
\begin{align}
\label{eq:cmixup}
P((x_j, y_j)|(x_i, y_i)) \propto \text{exp}(-\frac{d(y_i, y_j)}{b^2})
\end{align}
where $d(y_i, y_j)$ is the label distance and $b$ is the bandwidth of a kernel function. Since the values of a probability mass function sum to one, \cmixup{} normalizes the calculated values. 
The bandwidth is the key parameter to tune. As the bandwidth increases, the probability distribution becomes more uniform, thereby making \cmixup{} similar to Mixup. As the bandwidth decreases, samples are only mixed with their nearest neighbors, and \cmixup{} eventually becomes identical to Empirical Risk Minimization (ERM).

\paragraph{Robust Training}

There is an existing literature\,\cite{NEURIPS2018_a19744e2, jiang2018mentornet, chen2019understanding, DBLP:conf/icml/ShenS19,DBLP:conf/icml/SongK019,doi:10.1080/01621459.1984.10477105} on robust training where the goal is to perform accurate model training against noisy data. While there are many approaches, we focus on {\em multi-round robust training} where noisy data is either removed or fixed progressively during the model training iterations. 

As a default, we use Iterative Trimmed Loss Minimization (ITLM) \cite{DBLP:conf/icml/ShenS19} as a representative clean sample selection method, although we can use other methods as we demonstrate in Section~\ref{sec:otherrobust}. ITLM repeatedly removes a certain percentage of the training data where the intermediate model's predictions differ from the labels and solves the following optimization problem:
\begin{align*}
\min_{S:|S| = \lfloor\tau n\rfloor} \sum_{s_i \in S}{\ell_\theta(s_i)}\nonumber
\end{align*}
where $S \subseteq \train$ is the current clean data, $s_i$ is a sample in $S$, $\ell_\theta$ is a loss function, and $\tau$ is a parameter that indicates the ratio of clean samples in the training data. ITLM is widely used because of its simplicity and scalability. 

However, any other multi-round robust training technique\,\cite{DBLP:conf/icml/ChenLCZ19,DBLP:conf/iccv/HuangQJZ19,DBLP:conf/icml/SongK019} can be used as well. We demonstrate this point in Section~\ref{sec:otherrobust} where we replace ITLM with O2U-Net\,\cite{DBLP:conf/iccv/HuangQJZ19}, which is a different clean sample selection method, and SELFIE\,\cite{DBLP:conf/icml/SongK019}, which refurbishes labels of unclean samples.

\section{\cmixup{} Vulnerability to Noisy Data}

\cmixup{} assumes that all samples are clean data and performs Mixup using a fixed bandwidth. Although \cmixup{} is known to be robust against covariate shifts in the data where the distribution $P(X)$ may change, but not $P(Y|X)$, the data is still assumed to be clean. In comparison, we consider noisy data where $P(Y|X)$ may change as well.

\begin{figure}[t]
\centering
\begin{subfigure}{0.47\columnwidth}
\includegraphics[width=\columnwidth]{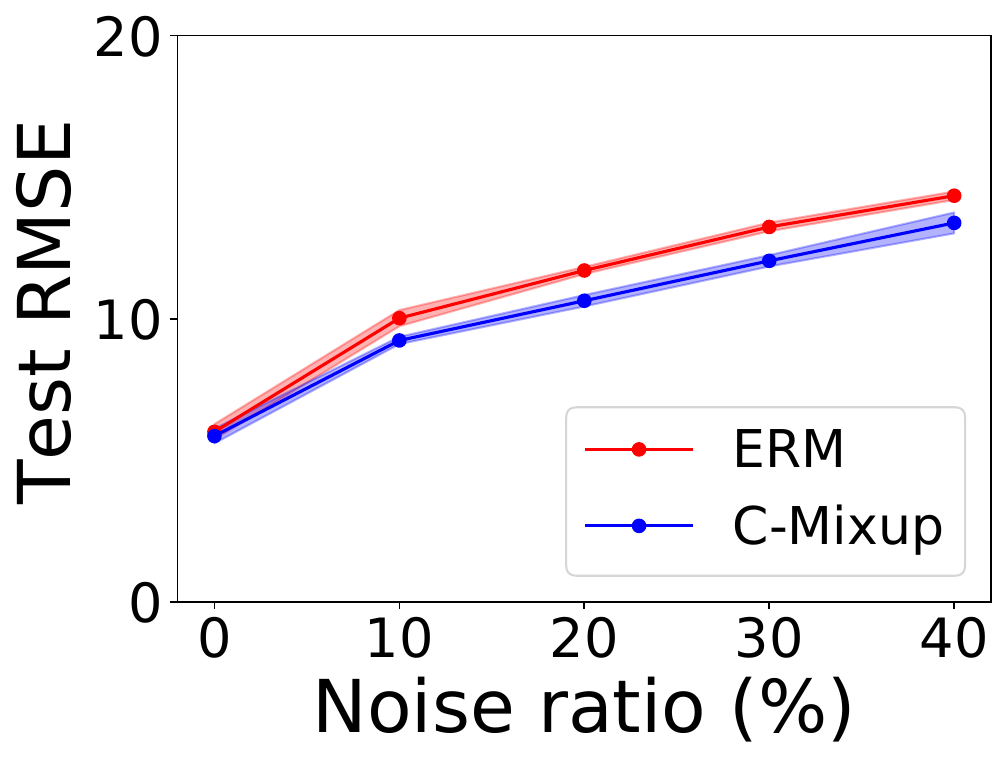}
\caption{}
\label{fig:vulnerable}
\end{subfigure}
\begin{subfigure}{0.47\columnwidth}
\includegraphics[width=\columnwidth]{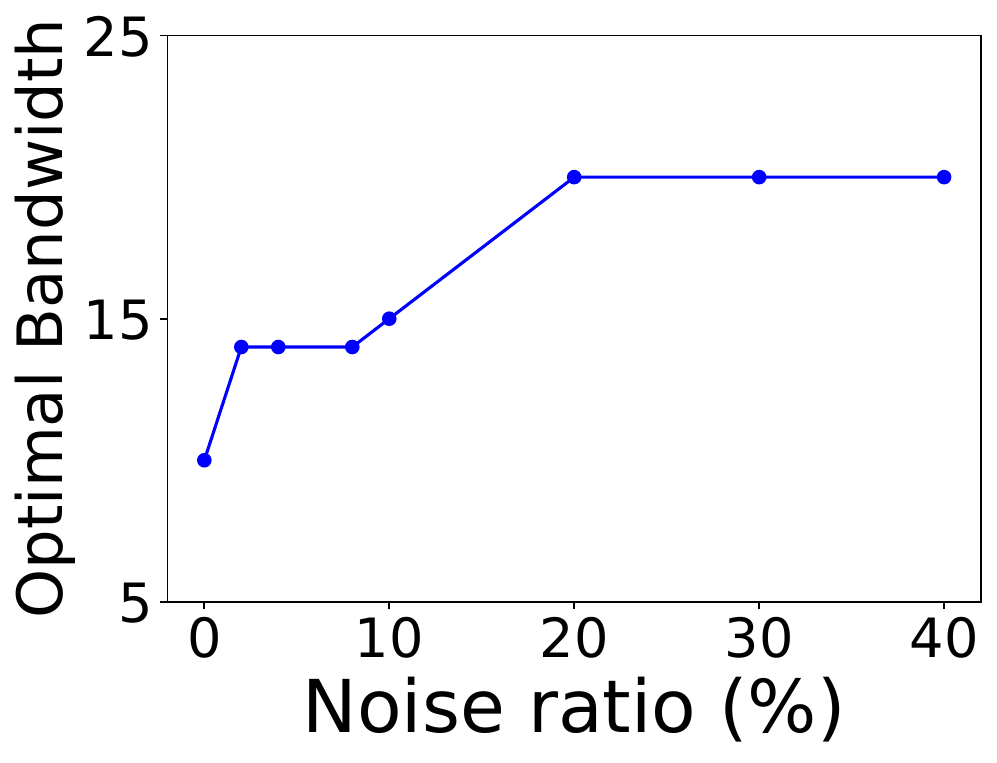}
\caption{}
\label{fig:noise_bw}
\end{subfigure}
\vskip -0.05in
\caption{We evaluate \cmixup{} on noisy data where we add label noise to the {\sf Spectrum} dataset. A lower RMSE means better model performance. (a) As the noise ratio increases, both ERM and \cmixup{} gradually perform worse where their performance gap does not change much. (b) In addition, the optimal bandwidth may actually increase where mixing dilutes out-of-distribution data from negatively impacting the model performance.}
% \label{fig:synergy}
\vskip -0.05in
\end{figure}

We demonstrate how \cmixup{} is vulnerable to such general noise in Figure~\ref{fig:vulnerable}. We experiment on the {\sf Spectrum} dataset for manufacturing and add random Gaussian noise to the labels (see more details in Section~\ref{sec:experiments}). We evaluate \cmixup{} with the tuned bandwidth on noisy data. This model is then evaluated on clean test data. We compare \cmixup{} with using ERM only and compare the Root Mean Squared Error (RMSE, see definition in Section~\ref{sec:experiments}) results where a lower value is better. As the noise ratio increases, the RMSE of \cmixup{} trained on noisy data increases. In addition, regardless of the noise ratio, the RMSE of \cmixup{} increases as much as ERM's RMSE does, showing the vulnerability of \cmixup{}.

To better understand how C-Mixup is affected by noise, we also determine the optimal bandwidth, which results in the best model performance across the different noise ratios as shown in Figure~\ref{fig:noise_bw} using the same experimental setup. As a result, as the noise ratio increases, the optimal bandwidth also increases. This result is counter-intuitive at first glance because it seems like noisy data should be mixed less. However, we suspect that mixing also has a dilution effect where out-of-distribution samples have less impact on the model training if they are mixed with other clean samples. We do not claim that this trend always holds, and the point is that mixing has a non-trivial effect on noisy data. Thus, it becomes difficult to find a single bandwidth that works best for both clean and noisy data.

The \cmixup{} paper\,\cite{DBLP:conf/nips/YaoWZZF22} also performs a robust training experiment against label noise, but it assumes a fixed noise ratio and does not necessarily show the entire story. The more extensive results here suggest that \cmixup{} is not designed to handle noise and can benefit from robust training methods. We thus propose a natural extension by integrating \cmixup{} with robust training.

\section{Problem Definition}

We formulate our problem as solving the bilevel optimization problem of cleaning and augmenting samples during model training:
\begin{gather}
 \min_{\theta} \sum_{s_i \in \mathtt{\cmixup{}}(S_c, b, \alpha)} \ell_\theta(s_i) \\
 \begin{aligned}[t]
        & S_c = \argmin_{\substack{S=\{s_i|p_i=1\}}} && \sum_{i=1}^{n}\ell_\theta(s_i) ~ p_i \\
        && \text{s.t.} & \sum_{i=1}^{n} p_i = \tau n \\
        &&& p_i \in \{0,1\}, i=1,...,n  
 \end{aligned}
\end{gather}
where $p_i$ indicates whether the data sample $s_i$ is selected or not, $\tau$ is the ratio of clean samples in the training data, $S$ is the selected samples, $\mathtt{\cmixup{}}(S)$ is an augmented result of $S$ by \cmixup{}, $b$ is the bandwidth, and $\alpha$ is the Mixup parameter. 

\section{\method{}}

We explain how we interleave data augmentation (\cmixup{}) with robust training, empirically analyze how the two techniques have a synergistic effect, propose bandwidth tuning techniques, and present the entire \method{} algorithm. 

\subsection{\cmixup{} and Robust Training Integration}

We solve the bilevel optimization problem by interleaving \cmixup{} with the robust training rounds. This approach is common in other bilevel optimization works \,\citep{liu2018darts, roh2021sample}. Recall that robust training iteratively cleans data and then updates its regression model until the data is clean enough. For each round, we can perform \cmixup{} between the data cleaning and model updating steps as in Figure~\ref{alg:rcmixup}. We explain how this sequence is beneficial to both methods.

\paragraph{Data Cleaning benefits \cmixup{}}

\begin{figure}[t]
\centering
\begin{subfigure}{0.47\columnwidth}
\includegraphics[width=\columnwidth]{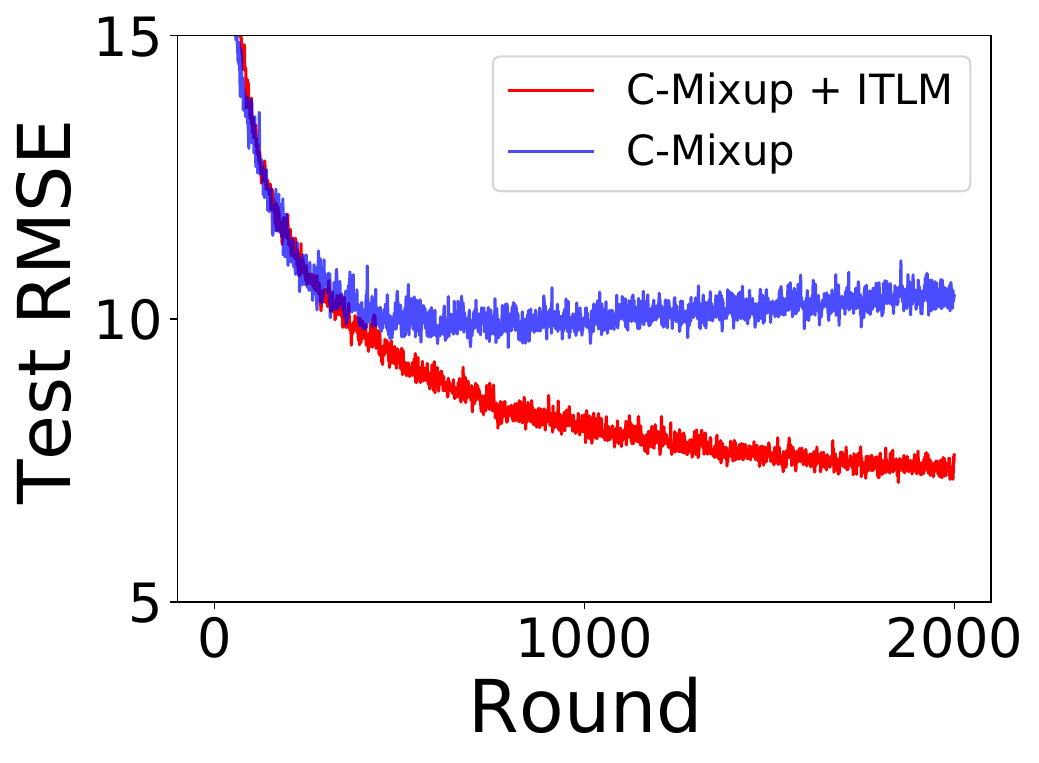}
\caption{}
\label{fig:synergy1}
\end{subfigure}
\begin{subfigure}{0.47\columnwidth}
\includegraphics[width=\columnwidth]{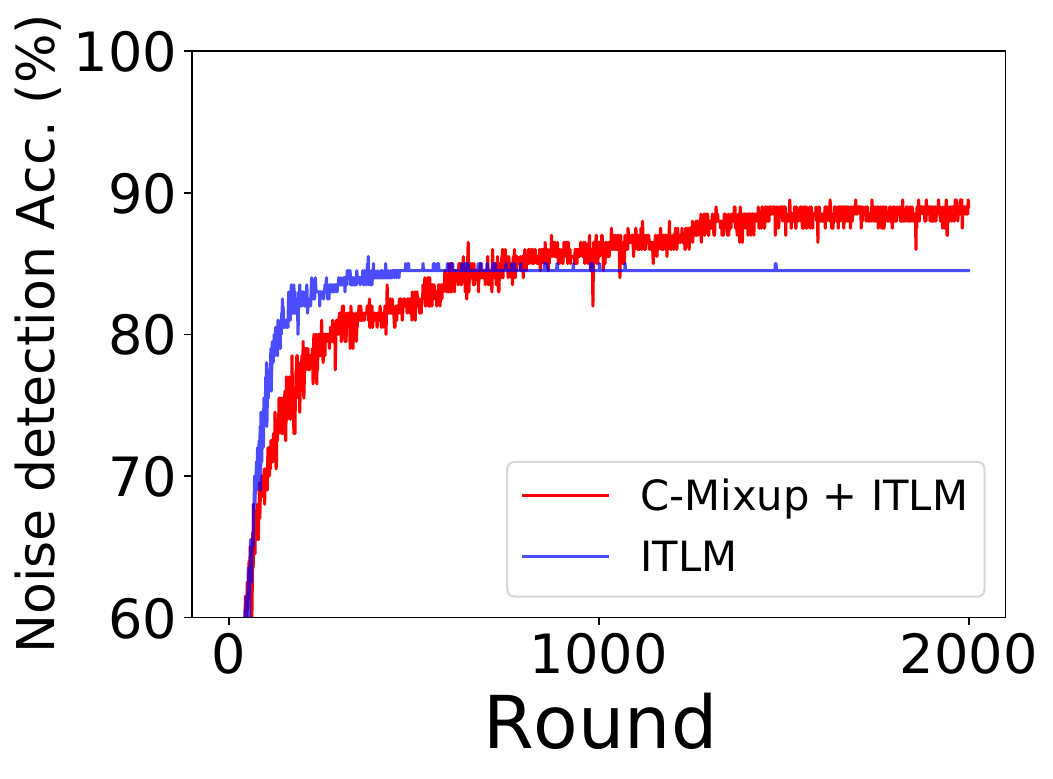}
\caption{}
\label{fig:synergy2}
\end{subfigure}
\vskip -0.05in
\caption{Robust training benefits \cmixup{} and vice versa. (a) As robust training iteratively cleans the data, \cmixup{}'s model performance improves. (b) Using \cmixup{} within robust training helps it remove noisy data better compared to when not using \cmixup{}.}
\vskip -0.1in
% \label{fig:synergy}
\end{figure}

We provide a simple empirical analysis of how data cleaning during robust training benefits \cmixup{} in Figure~\ref{fig:synergy1}. We experiment on the {\sf Spectrum} dataset used above and add random Gaussian noise to the labels, independently per label. We start with a 10\% noise ratio and run ITLM, which progressively cleans the data. For each round, we evaluate \cmixup{} by training a model on the mixed data. As a result, when combining \cmixup{} with robust training, the model's RMSE on the validation set clearly improves with each round, unlike when using \cmixup{} only. In addition, while using \cmixup{} only has a slight performance decrease after convergence due to overfitting to the noisy data, \cmixup{} with robust training does not have this problem.

\paragraph{\cmixup{} benefits Data Cleaning Performance}

We also empirically analyze how \cmixup{} improves robust training by making it identify clean data better in Figure~\ref{fig:synergy2}. We use the same experimental setup as above and evaluate the noise detection accuracy of ITLM, which is the percent of noisy samples that are correctly identified by robust training. We compare ITLM with when it is combined with \cmixup{}. As a result, \cmixup{} improves ITLM's noise detection accuracy by up to 5\%.

\subsection{Dynamic Bandwidth Tuning}

As we analyzed in Figure~\ref{fig:noise_bw}, the optimal bandwidth of \cmixup{} may vary as the data is progressively cleaned via robust training, so we would like to dynamically adjust this value both accurately and efficiently. Recall that the bandwidth adjusts the level of mixing where a larger bandwidth means that samples are mixed more with neighbors. Following the original \cmixup{} setup\,\cite{DBLP:conf/nips/YaoWZZF22}, we consider a fixed set of bandwidth candidates $B = \{b_1, b_2, \ldots, b_{|B|}\}$ and aim to select the best one. 

Our strategy is to update the best bandwidth after every $L > 0$ rounds of robust training. In the initial warm-up phase, we opt for a higher bandwidth to accommodate the relatively greater noise ratio. Subsequently, for each update, we evaluate bandwidth candidates within $B$ on the next $N > 0$ rounds and then choose the one that results in the best model performance. Analytically predicting model performance after multiple rounds is challenging because the cleaned data itself keeps on changing. 
While this strategy has an overhead, we later show in  Section~\ref{sec:parameteranalysis} that only one or two initial updates are needed to achieve most of the performance benefits.

One way to further reduce the overhead is to avoid the bandwidth searching altogether and instead simply decay the bandwidth by a certain ratio (say by 10\%) every $L$ epochs based on the observation that the optimal bandwidth tends to decrease for lower noise ratios. Naturally there is be a tradeoff between runtime and performance, which we show in Section~\ref{sec:modelperformanceandruntime}. In addition, we now need to figure out the right decay rate. Nonetheless, if lowering runtime is critical, the decaying strategy can be a viable option.

\subsection{Overall Algorithm}

Algorithm~\ref{alg:rcmixup} shows the overall \method{} algorithm when using ITLM as the robust training method. 
We first initialize model parameters using C-Mixup (Steps 2--4). The initial bandwidth is tuned on the validation set. This bootstrapping is important because  the later steps are designed to gradually update its value as the data is cleaned progressively, so we need to start from a bandwidth value that is optimal on the noisy data first. Next, for each round, robust training cleans the data, we run \cmixup{} and update the model on the cleaned data (Steps 18--20). After every $L$ rounds, we also update the bandwidth of \cmixup{} by evaluating the possible bandwidths on $N$ rounds and choosing the one that results in the lowest validation set RMSE (Steps 8--16). We repeat the entire process until the model converges. Algorithm~\ref{alg:clean} shows the clean sample selection algorithm in ITLM, and Algorithm~\ref{alg:cmixup} shows the C-Mixup algorithm.

\SetKwComment{Comment}{// }{}
\begin{algorithm}[t]
\SetKwInput{Input}{Input}
\SetKwInOut{Output}{Output}
\SetNoFillComment
%\small
\Input{Training data $\train$, validation set $\train^v$, 
%number of rounds $R$, 
bandwidth update interval $L$, 
%number of neighboring bandwidths $K$ to search, 
number of bandwidth update rounds $N$, possible bandwidths $B = \{b_1, \ldots, b_{|B|}\}$, clean ratio $\tau$, Mixup parameter $\alpha$, initial bandwidth $b$, initial model parameters $\theta$}
\Comment{Warm-up phase}
\For {$\mathrm{round=1\ to\ }$warm-up rounds} {
$S_{mix} = \texttt{\cmixup{}} (\train, b, \alpha)$\\
Update model parameters $\theta$ on $S_{mix}$ \\
}
$i \leftarrow 0$\\
\Comment{Clean \& Update phase}
\While {$\mathrm{not\ converge}$}{
    \If {$i~\%~L$ == 0} {
        \Comment{Update bandwidth}
        \For {$b_j \mathrm{\ in\ } B$} {
        $\theta_{b_j} \leftarrow \theta$ \\
          \For {$\mathrm{round=1\ to\ } N$} {
            $C = \texttt{CleanSelection} (\train, \theta_{b_j}, \tau)$\\
            $S_{mix} = \texttt{\cmixup{}} (C, b_j, \alpha)$\\
            Update model parameters $\theta_{b_j}$ on $S_{mix}$ \\
          }
      }
        $(b^*, \theta_{b^*})$ = Bandwidth and corresponding model parameters with the lowest RMSE\\
        $\theta \leftarrow \theta_{b^*}$ \\
        $b \leftarrow b^*$ \\
    } \Else {
        $C = \texttt{CleanSelection} (\train, \theta, \tau)$\\
        $S_{mix} = \texttt{\cmixup{}} (C, b, \alpha)$\\
        Update model parameters $\theta$ on $S_{mix}$ \\
    }
    $i = i + 1$\\
}
\Output{$\theta$}
\caption{The \method{} algorithm.}
\label{alg:rcmixup}
\end{algorithm}

\begin{algorithm}[t]
\SetKwInput{Input}{Input}
\SetKwInOut{Output}{Output}
\SetNoFillComment
%\small
\Input{Dataset $D$, model parameters $\theta$, clean ratio $\tau$}
%\text{profit}$ = $\max(\ell_\theta(\train))-\ell_\theta(\train)$\\
%$\text{sortIdx}$ = $\texttt{argSort}(\text{profit})$ \\
$\text{sortIdx}$ = $\texttt{argSort}(\ell_\theta(D))$ (ascending order) \\
$C$ $\gets$ first $\tau |D|$ elements of $D[\text{sortIdx}]$\\
\Output{$C$}
\caption{The clean sample selection algorithm.}
\label{alg:clean}
\end{algorithm}

\begin{algorithm}[t]
\SetKwInput{Input}{Input}
\SetKwInOut{Output}{Output}
\SetNoFillComment
%\small
\Input{Dataset $D$, bandwidth $b$, Mixup parameter~$\alpha$}
$S \leftarrow []$\\
\For{$(x_i, y_i) \mathrm{\ in\ } D$}{
Sample $(x_j, y_j)$ using Equation~\ref{eq:cmixup} and $b$\\
Sample $\lambda \sim Beta(\alpha, \alpha)$\\
$\tilde{x} = \lambda x_i + (1 - \lambda) x_j$\\
$\tilde{y} = \lambda y_i + (1 - \lambda) y_j$\\    
$S \leftarrow S \cup \{(\tilde{x}, \tilde{y})\}$\\
}
\Output{$S$}
\caption{The \cmixup{} algorithm.}
\label{alg:cmixup}
\end{algorithm}

\paragraph{Overhead on Robust Training}

In comparison to robust training, \method{} adds the overhead of mixing samples via \cmixup{} during each round. For every $L$ rounds, there is also the overhead of tuning the bandwidth using $N$ rounds of training. In our experiments, we observe that the tuning overhead is small because the bandwidth only needs to be updated once or twice.

\section{Experiments}
\label{sec:experiments}

We provide experimental results for \method{}. We evaluate the regression models trained on the augmented training sets on separate test sets. We use Root Mean Squared Error (RMSE) and Mean Absolute Percentage Error (MAPE) for measuring model performance where lower values are better. We report the mean and the standard deviation ($\pm$ in tables) for results of five random seeds. We use PyTorch\,\citep{paszke2017automatic}, and all experiments are performed using Intel Xeon Silver 4210R CPUs and NVIDIA Quadro RTX 8000 GPUs. %More setting details are in the appendix. 

\paragraph{Metrics}
RMSE is defined as $\sqrt{\frac{1}{n}\sum^n_{i=1}(y_i-\hat{y}_i)^2}$. MAPE is defined as $\frac{1}{n}\sum^n_{i=1}\left\lvert\frac{y_i-\hat{y}_i}{y_i}\right\lvert \times 100$ and is a measure of prediction performance of forecasting methods.

\paragraph{Experimental environment}
We conducted all experiments using Intel(R) Xeon(R) Silver 4210R CPUs @ 2.40GHz, and eight NVIDIA Quadro RTX 8000 GPUs equipped with 48GB VRAM on Linux Ubuntu 18.04.5. We employed Python 3.8.13, Scikit-learn version 0.24.2, and PyTorch version 1.7.1 for our evaluations.

\paragraph{Datasets}
We use one synthetic and three real datasets. The {\sf Spectrum} dataset\,\citep{dacon} is synthetic and contains spectrum data generated by applying light waves on 4-layer 3D semiconductors and measuring the returning wavelengths. This data is used to predict layer thickness without touching the semiconductor itself. The {\sf NO2} emissions dataset\,\citep{statlib} contains traffic and meteorological information around roads and is used to predict NO2 concentration. The {\sf Airfoil} dataset\,\cite{Dua:2019} contains aerodynamic and acoustic test results for airfoil blade sections in a wind tunnel. Finally, the {\sf Exchange-Rate}\,\citep{lai2018modeling} is a time-series dataset and contains daily exchange rates from 8 countries. The three real datasets are from \cite{DBLP:conf/nips/YaoWZZF22} for a fair comparison. Table~\ref{tbl:datasetcomparison} shows the dimension and size information of the datasets.

\begin{table}[t]
 \vskip -0.07in
  \setlength{\tabcolsep}{4pt}
  \caption{Settings for the four datasets.}
  \centering
  %\small
  \begin{tabular}{ccccc}
    \toprule
    Dataset & Data dim. & Label dim. & $|\train|$ & $|\train^v|$\\
    \midrule
    {\sf Spectrum} & 226 & 4 & 2,000 & 500\\
    {\sf NO2} & 7 & 1 & 200 & 200 \\
    {\sf Airfoil} & 5 & 1 & 1,000 & 400\\
    {\sf Exchange-Rate} & 168$\times$8 & 8 & 4,373 & 1,518\\
%    {\sf Product} & 580 & 20 & 300 & 200 \\
    % {\sf Bike} & 12 & 1 & 300 & 200 \\
    \bottomrule
  \end{tabular}
  \label{tbl:datasetcomparison}
  \vskip -0.12in
\end{table}

\paragraph{Noise Injection}
We inject noise to labels using two methods: (1) Gaussian noise, which adds to each label a value sampled from the Gaussian distribution $N(0, m^2\sigma^2)$, where $m$ is the noise magnitude (see default values in the appendix), and $\sigma$ is the standard deviation of labels of training set and (2) labeling flipping noise, which subtracts a maximum label value by each label as in classification\,\cite{DBLP:conf/pkdd/PaudiceML18}. We use a noise rate of 10--40\% where the default is 30\%. We do not set the noise ratio to be larger than 50\% to prevent the noise from dominating the data.

\paragraph{Baselines}
We consider two types of baselines:
\begin{itemize}
    \item {\em Individual methods}: performing ITLM\,\cite{DBLP:conf/icml/ShenS19} only (``Rob. training'') and performing \cmixup{}\,\cite{DBLP:conf/nips/YaoWZZF22} only (``\cmixup{}'').
    \item {\em Simple combinations}: performing robust training first and \cmixup{} in sequence (``R$\rightarrow$C''), performing \cmixup{} and then robust training in sequence (``C$\rightarrow$R''), and performing \method{} without dynamic bandwidth tuning (``C$\rightarrow$R+C'').
\end{itemize}

\paragraph{Parameters} 
For robust training, we assume that the clean ratio $\tau$ is known for each dataset. If $\tau$ is unknown, it can be inferred with cross validation~\citep{liu2015classification, yu2018efficient}. For the \cmixup{}, C$\rightarrow$R, and C$\rightarrow$R+C baselines, we use a single bandwidth value and tune it using a grid search. We choose other parameters using a grid search on the validation set or following \cmixup{}'s setup. More details on parameters are in the appendix.

\begin{table}[t]
  \setlength{\tabcolsep}{4pt}
  \centering
  %\small
  \caption{\method{} performance compared to the five baselines on the real and synthetic datasets. }
  \begin{tabular}{clll}
  \toprule
    Dataset & Method & RMSE & MAPE \\
    \midrule
    \multirow{7}{*}{\sf Spectrum} & {\cmixup{}} & {$12.125_{\pm 0.200}$} & {$10.840_{\pm 0.246}$} \\% & 500\\
    & {Rob. training} & {$9.756_{\pm 0.541}$} & {$7.228_{\pm 0.221}$} \\% & 666\\
    \cmidrule(lr){2-4}
    & R$\rightarrow$C & {$9.055_{\pm 0.445}$} & {$6.806_{\pm 0.335}$}  \\
    & C$\rightarrow$R & {$8.024_{\pm 0.321}$} & {$6.468_{\pm 0.211}$}  \\
    & C$\rightarrow$R+C & {$8.755_{\pm 0.187}$} & {$7.031_{\pm 0.067}$}  \\
    \cmidrule(lr){2-4}
    & \method{} & {$\textbf{7.471}_{\pm 0.220}$} & {$\textbf{5.930}_{\pm 0.165}$}\\
    \midrule
    \multirow{7}{*}{\sf NO2} & {\cmixup{}} & {$0.586_{\pm 0.033}$} & {$14.604_{\pm 1.283}$}\\
    & {Rob. training} & {$0.574_{\pm0.028}$} & {$14.418_{\pm 1.551}$} \\
    \cmidrule(lr){2-4}
    & R$\rightarrow$C & {$0.573_{\pm 0.023}$} & {$14.340_{\pm 1.258}$} \\
    & C$\rightarrow$R & {$0.562_{\pm 0.038}$} & {$14.020_{\pm 1.562}$}  \\
    & C$\rightarrow$R+C & {$0.566_{\pm0.033}$} & {$14.203_{\pm 1.549}$}  \\
    \cmidrule(lr){2-4}
    & \method{} & {$\textbf{0.557}_{\pm0.034}$} & {$\textbf{13.816}_{\pm 1.440}$} \\
    \midrule
    \multirow{7}{*}{\sf Airfoil} & {\cmixup{}} & {$3.438_{\pm0.218}$} & $2.093_{\pm 0.185}$  \\
    & {Rob. training} & {$2.760_{\pm 0.329}$} & $1.529_{\pm 0.069}$  \\
    \cmidrule(lr){2-4}
    & R$\rightarrow$C & {$3.226_{\pm 0.260}$} & {$1.833_{\pm 0.136}$}  \\
    & C$\rightarrow$R & {$2.721_{\pm 0.463}$} & {$1.492_{\pm 0.235}$}  \\
    & C$\rightarrow$R+C & {$2.699_{\pm0.381}$} & {$1.501_{\pm0.167}$}  \\
    \cmidrule(lr){2-4}
    & \method{} & {$\textbf{2.530}_{\pm0.357}$} & {$\textbf{1.398}_{\pm0.123}$}  \\
    \midrule
    \multirow{7}{16mm}{\sf Exchange-Rate} & {\cmixup{}} & {$0.0216_{\pm 0.0018}$} & {$2.2931_{\pm0.2153}$}  \\
    & {Rob. training} & {$0.0180_{\pm0.0008}$} & {$1.7715_{\pm0.1213}$} \\
    \cmidrule(lr){2-4}
    & R$\rightarrow$C & {$0.0165_{\pm 0.0015}$} & {$1.5825_{\pm0.1821}$} \\
    & C$\rightarrow$R & {$0.0162_{\pm 0.0011}$} & {$1.5553_{\pm0.1428}$} \\
    & C$\rightarrow$R+C & {$0.0162_{\pm 0.0011}$} & {$1.5495_{\pm0.1214}$} \\
    \cmidrule(lr){2-4}
    & \method{} & {$\textbf{0.0156}_{\pm0.0007}$} & {$\textbf{1.4692}_{\pm0.0811}$}  \\
  \bottomrule
  \end{tabular}
  \label{tbl:regressionaccuracy}
  \vskip -0.05in
\end{table}

\begin{figure*}[t]
\centering
    \subfloat[{\sf Spectrum}]{
        {\includegraphics[scale=0.22]{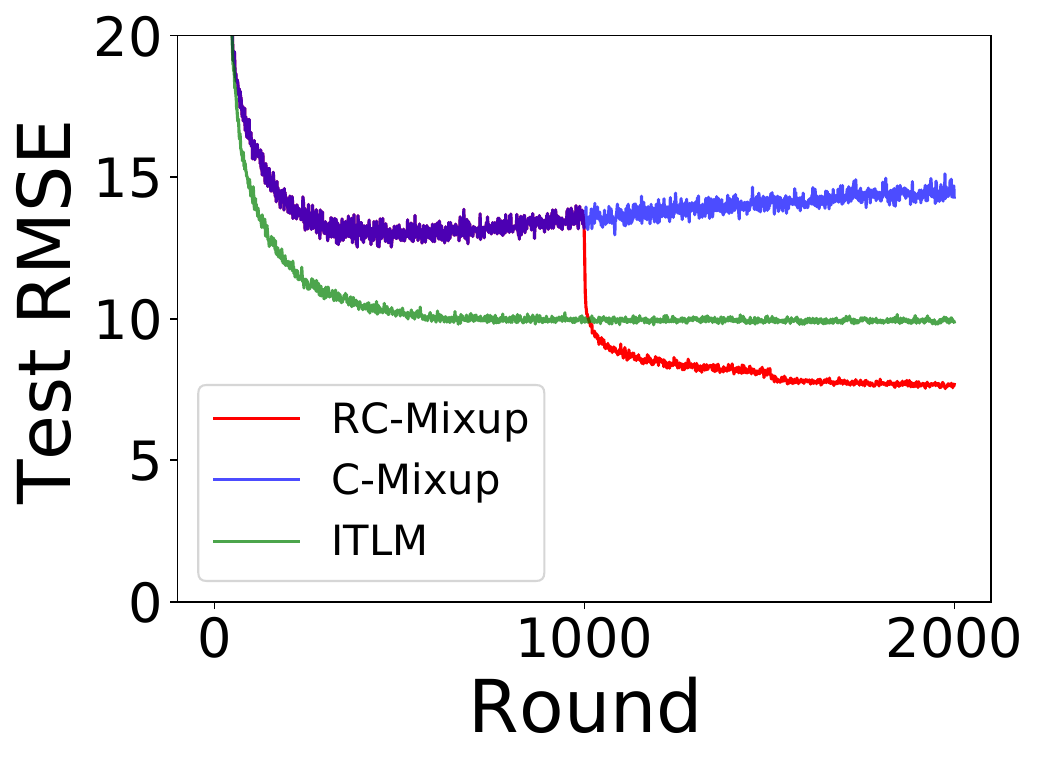}}
        \label{fig:convergence_spectrum}
        }
    \hspace{0.2cm}
    \subfloat[{\sf NO2}]{
        {\includegraphics[scale=0.22]{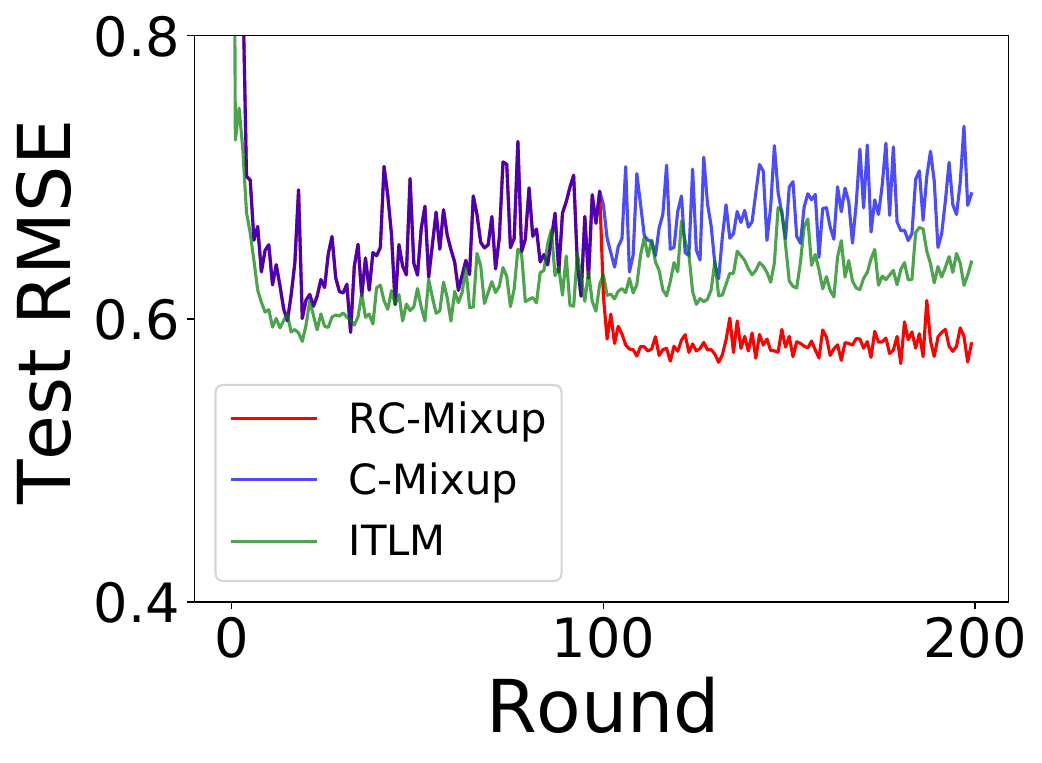}}
        \label{fig:bandwidth}
        }  
    \hspace{0.2cm}
    \subfloat[{\sf Airfoil}]{
        {\includegraphics[scale=0.22]{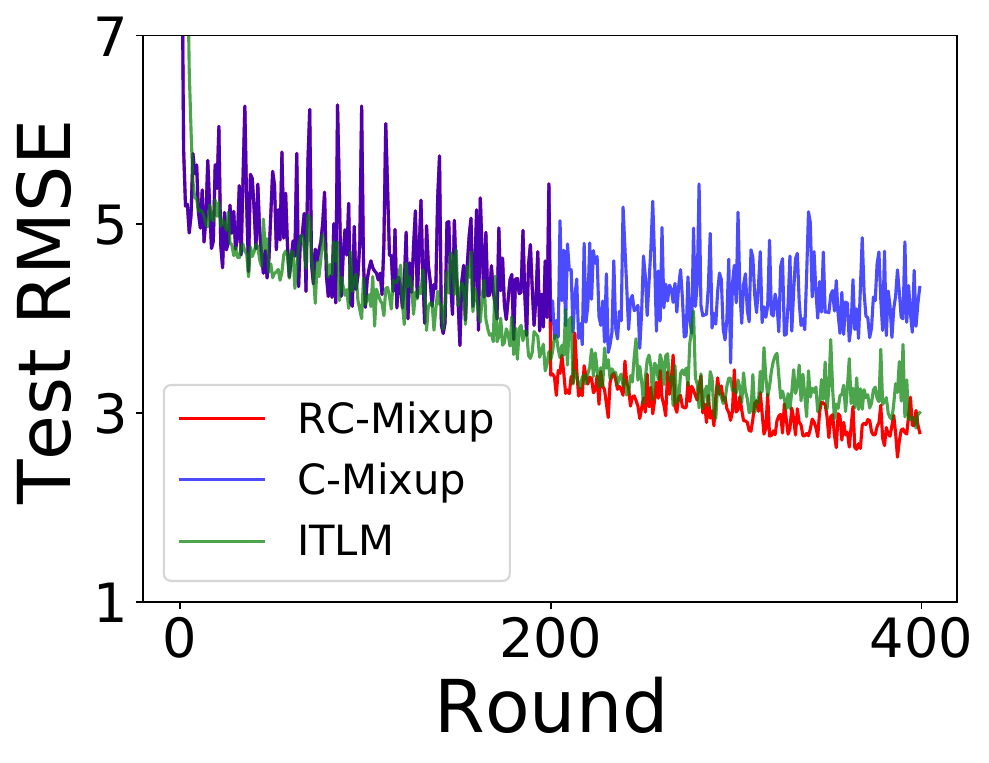}}
        \label{fig:convergence_airfoil}
        }
    \hspace{0.2cm}
    \subfloat[{\sf Exchange-Rate}]{
        {\includegraphics[scale=0.22]{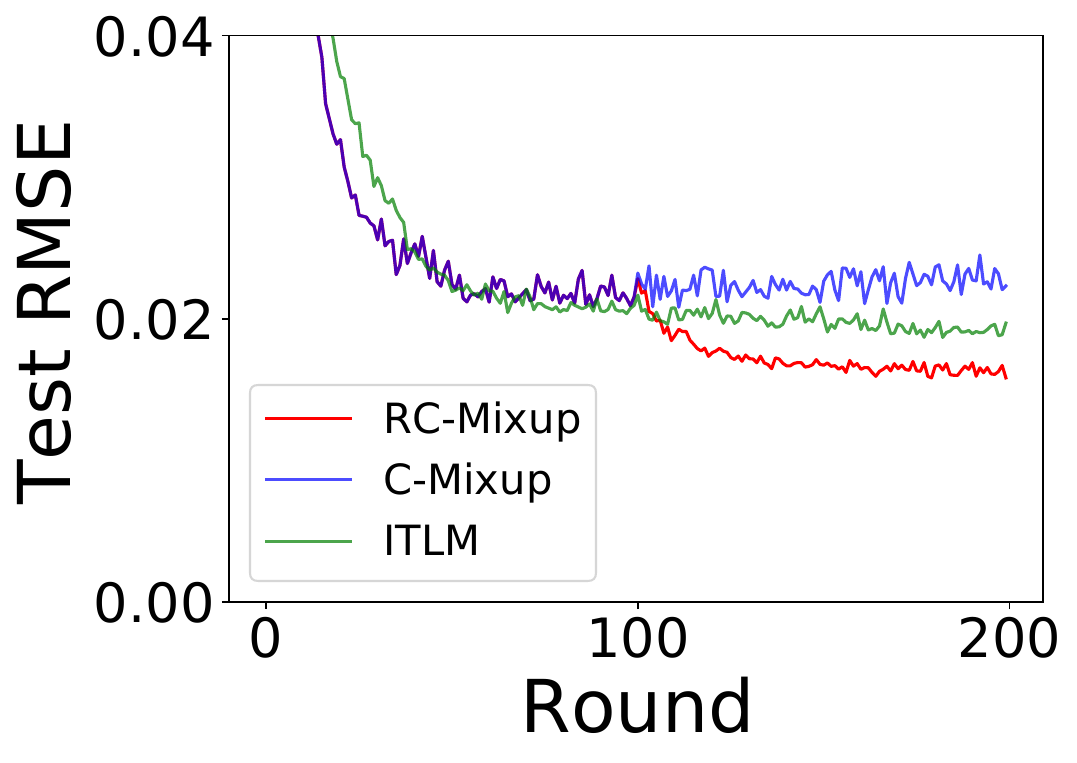}}
        \label{fig:convergence_exchange}
        }      \\
    \subfloat[{\sf Spectrum}]{
        {\includegraphics[scale=0.23]{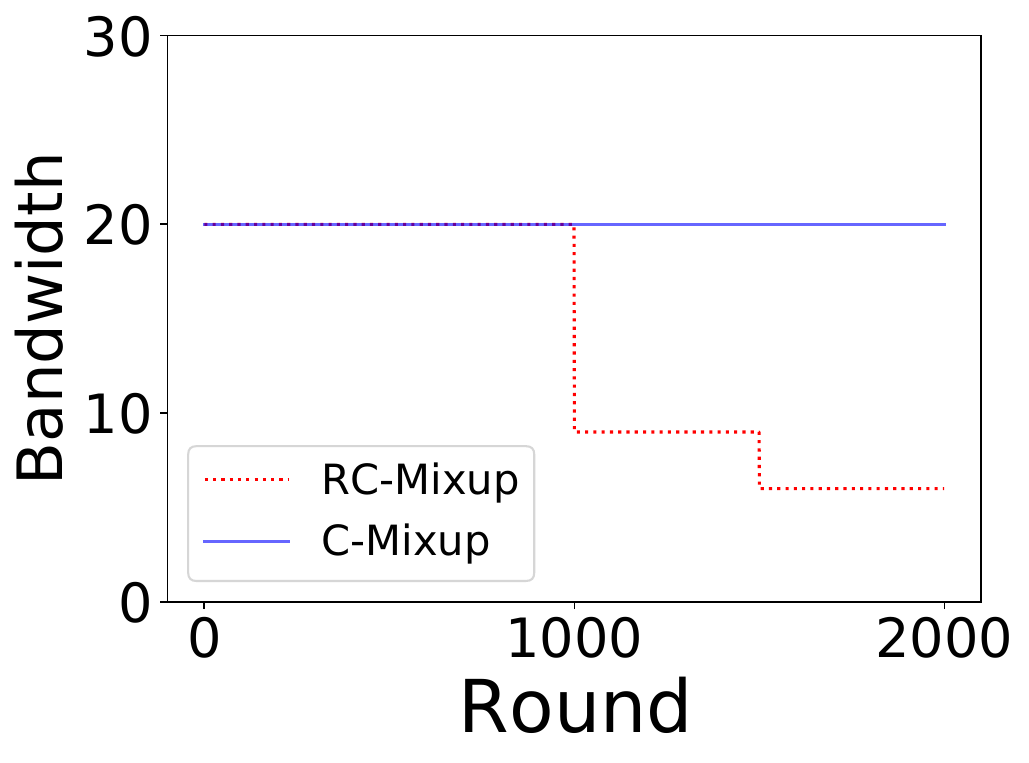}}
        \label{fig:bandwidth_spectrum}
        }
    \subfloat[{\sf NO2}]{
        {\includegraphics[scale=0.23]{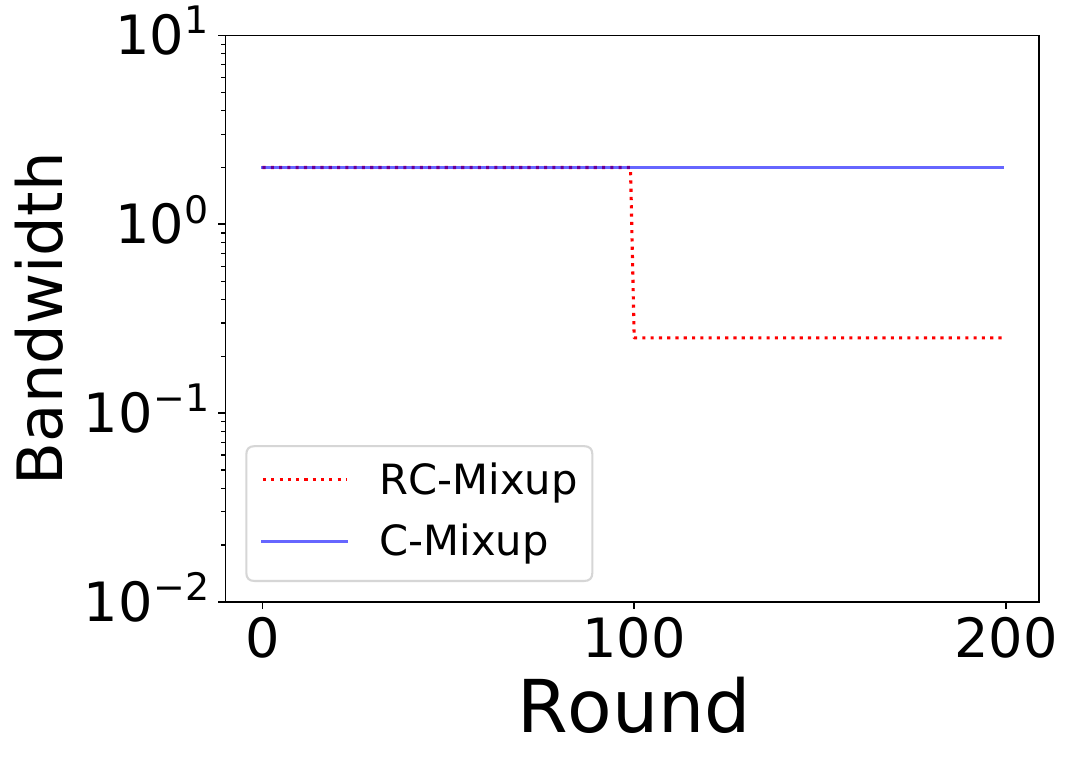}}
        \label{fig:bandwidth_no2}
        }
    \subfloat[{\sf Airfoil}]{
        {\includegraphics[scale=0.23]{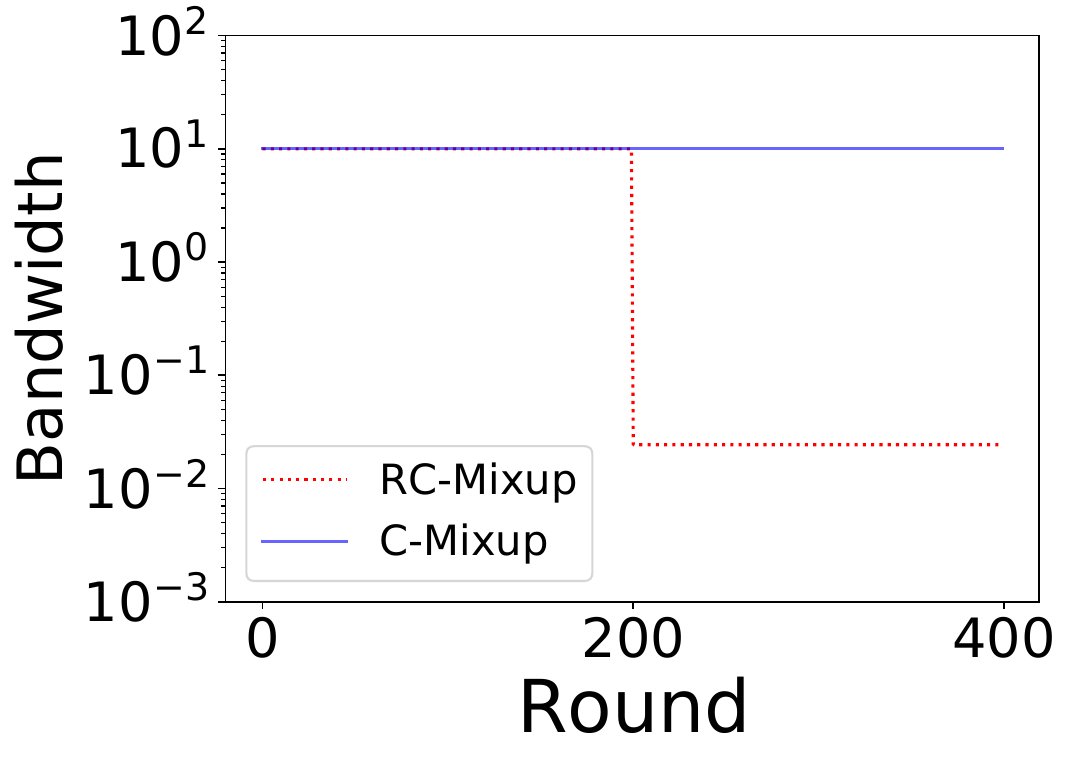}}
        \label{fig:bandwidth_airfoil}
        }
    \subfloat[{\sf Exchange-Rate}]{
        {\includegraphics[scale=0.23]{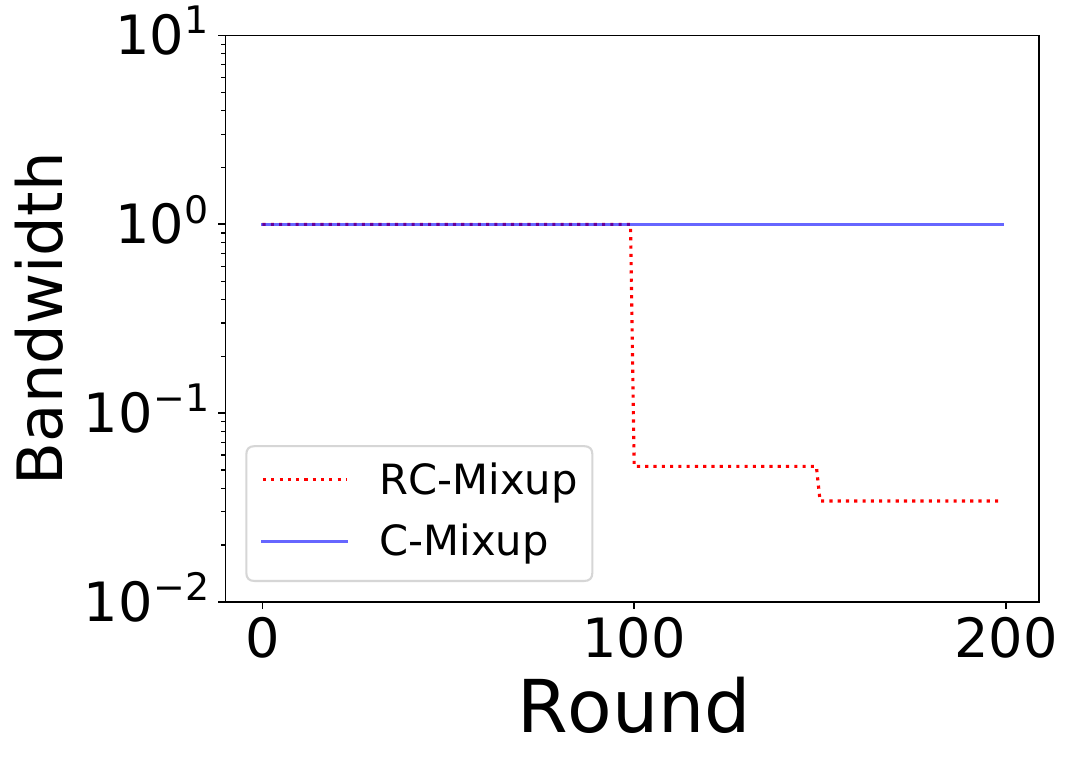}}
        \label{fig:bandwidth_exchange}
        } \\
    \subfloat[{\sf Spectrum}]{
        {\includegraphics[scale=0.22]{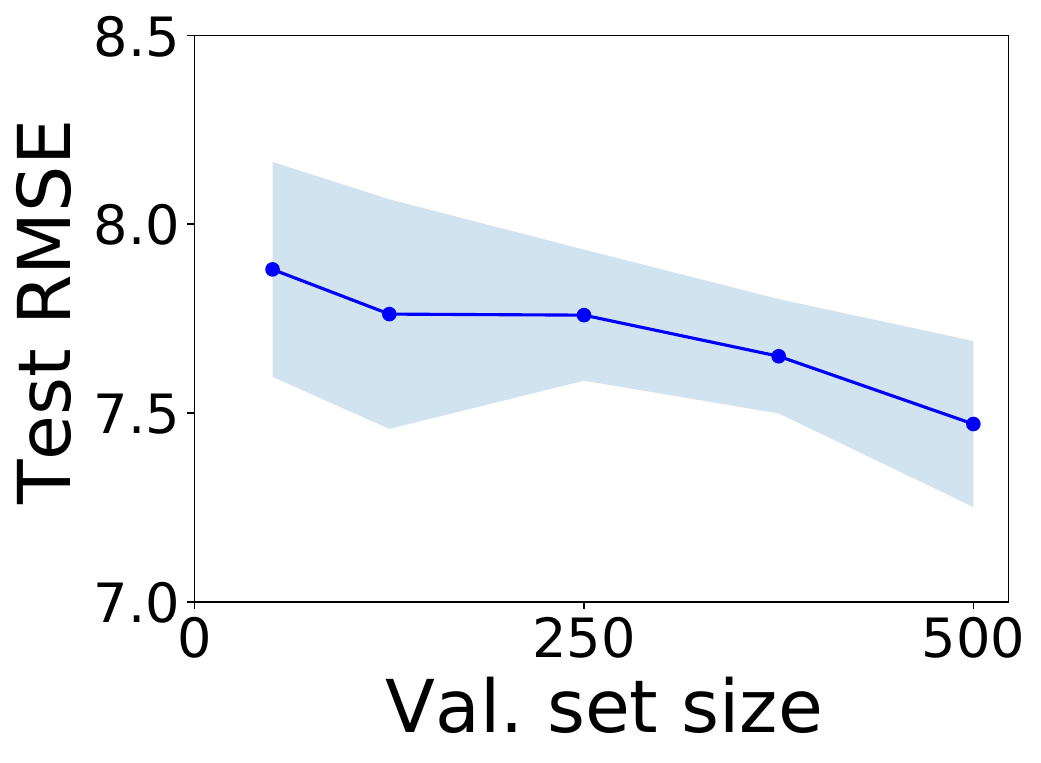}}
        \label{fig:valsize_spectrum}
        }
    \hspace{0.2cm}
    \subfloat[{\sf NO2}]{
        {\includegraphics[scale=0.22]{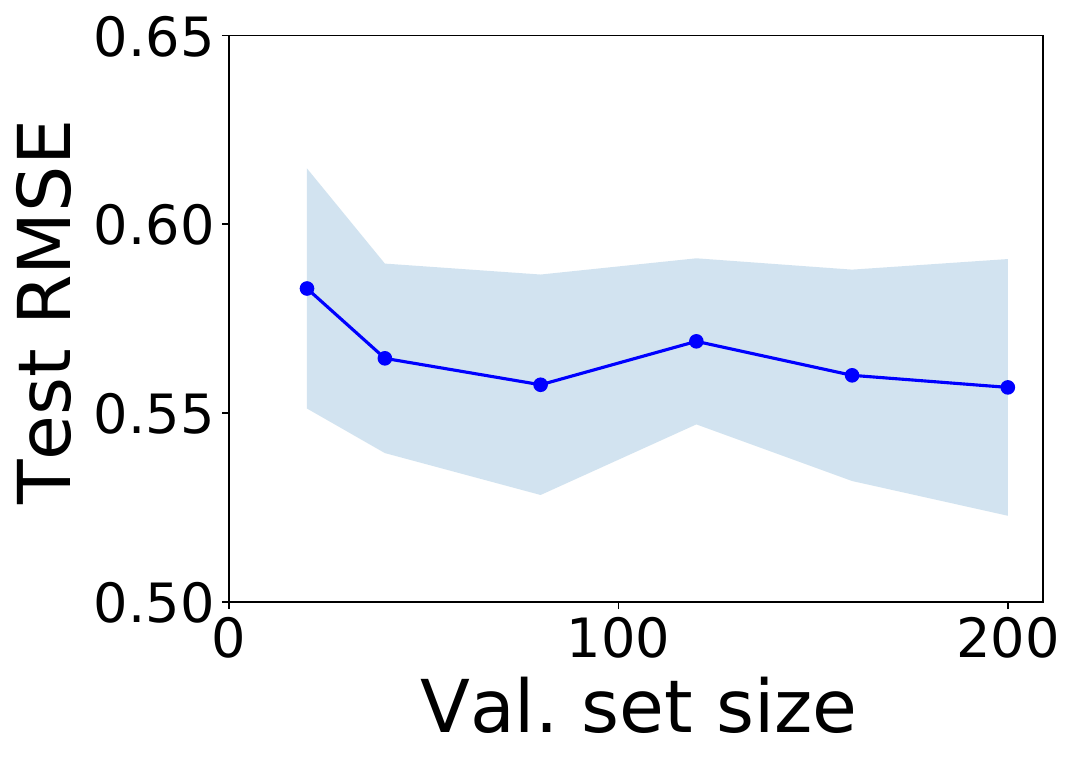}}
        \label{fig:valsize_spectrum}
        }
    \hspace{0.2cm}
    \subfloat[{\sf Airfoil}]{
        {\includegraphics[scale=0.22]{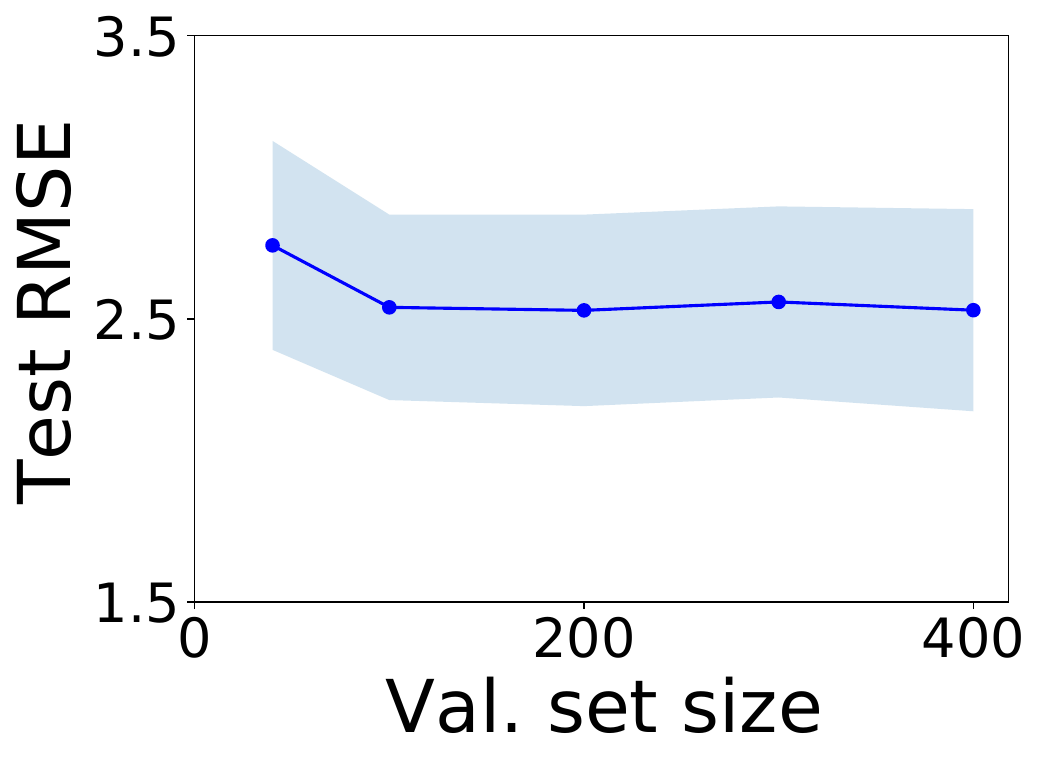}}
        \label{fig:valsize_spectrum}
        }
    \hspace{0.2cm}
    \subfloat[{\sf Exchange-Rate}]{
        {\includegraphics[scale=0.22]{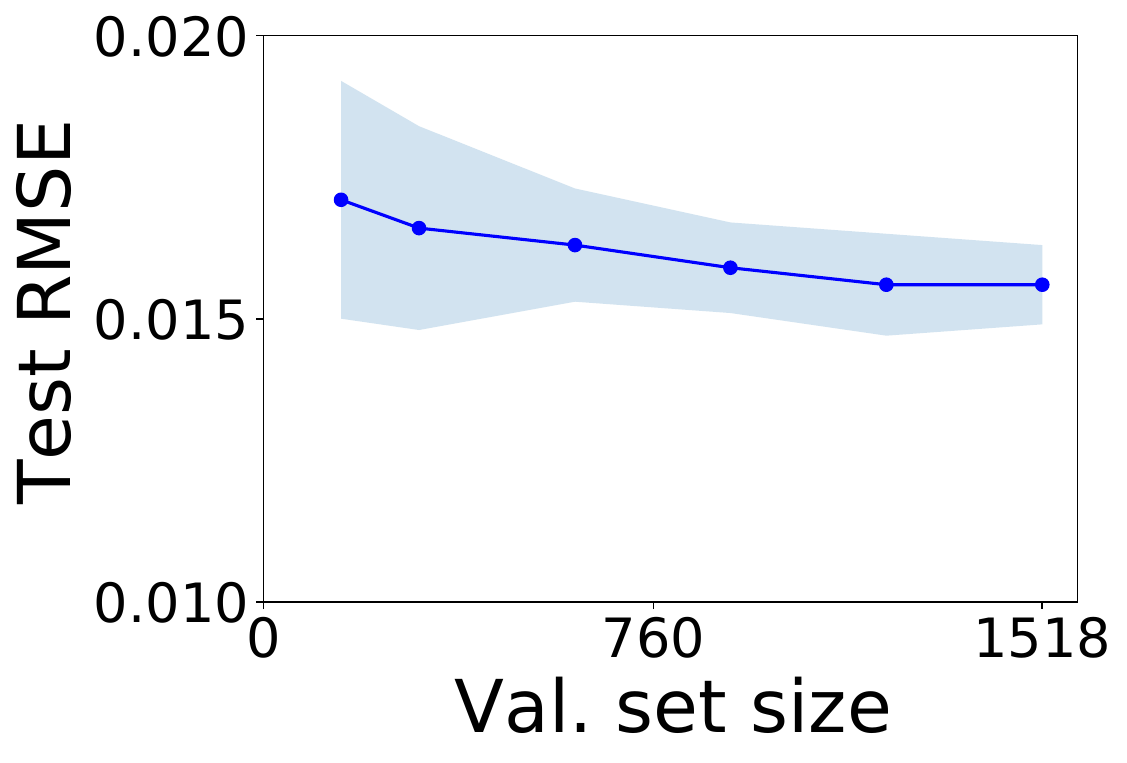}}
        \label{fig:valsize_spectrum}
        }
\vskip -0.1in
\caption{(a)-(d) \method{} model training convergence results. (e)-(h) \method{} dynamic bandwidth tuning results. The tuned bandwidth values vary slightly depending on the random seed, and we show the bandwidth averaged across five random seeds (dotted red lines). (i)-(l) \method{} performance results while varying the validation set size.}
\vskip -0.05in
\label{fig:convergence_bandwidth_validationsetsize}
\end{figure*}

\subsection{Model Performance and Runtime Results}
\label{sec:modelperformanceandruntime}

We compare the overall performance \method{} with the two types of baselines on all the datasets as shown in Table~\ref{tbl:regressionaccuracy}. For the individual method baselines, we observe that \cmixup{} is indeed vulnerable to noise, and that robust training is less affected by the noise, but still needs improvement. The simple combination baselines C$\rightarrow$R and C$\rightarrow$R+C increasingly outperform the individual methods as they start to take advantage of both methods. However, \method{} performs the best by also dynamically tuning the bandwidth. We also note that our results are near-optimal. As a reference, the (RMSE, MAPE) results in an optimal setting where we only use clean data are: (6.961, 6.000) for {\sf Spectrum} and (0.535, 13.209) for {\sf NO2}. The RC-Mixup results are similar to these results and cannot be further improved.

We also show how \method{}'s model training converges and how its bandwidth is tuned in Figures~\ref{fig:convergence_bandwidth_validationsetsize}a--\ref{fig:convergence_bandwidth_validationsetsize}d. For all four datasets, \method{} clearly converges to lower RMSE values for the same number of rounds compared to \cmixup{} and robust training. \method{}'s RMSE values drastically decrease around the middle of each figure because robust training and bandwidth tuning are applied at that point. Once robust training is applied, the model training is now done on cleaned data, so the model performance improves significantly afterwards. 
We then show how \method{} dynamically adjusts its bandwidth for the four datasets with five random seeds in Figures~\ref{fig:convergence_bandwidth_validationsetsize}e--\ref{fig:convergence_bandwidth_validationsetsize}h. For these datasets, the bandwidth values vary slightly depending on the random seed, and we show the bandwidth averaged across five random seeds (dotted red lines).
The decreasing trends of the tuned bandwidths are consistent with Figure~\ref{fig:noise_bw}, where as the data is cleaned, a smaller bandwidth is more effective. We do not claim this trend always holds, and the point is that \method{} is able to find the right bandwidth in any situation.

\paragraph{Runtime}

The computational cost of \method{} varies depending on the number of bandwidth candidates or bandwidth update frequency. For the {\sf Spectrum} dataset, the runtimes of all methods are as follows: robust training (ITLM) only: 52s, C-Mixup only: 244s, C$\xrightarrow{}$R: 161s, C$\xrightarrow{}$C+R: 250s, and RC-Mixup: 578s. Although RC-Mixup requires roughly twice the runtime of C-Mixup, the substantial performance improvement of RC-Mixup justifies its overhead. In general, while the efficiency decreases as the search space grows, we can obtain performance improvements with only 4--7 bandwidth candidates in practice. As a reference, \cmixup{} also searches for the optimal bandwidth among 6--10 bandwidth candidates using a grid search.

\paragraph{Bandwidth Decaying Results}
We also show \method{}'s performance and runtime when using bandwidth decaying in Table~\ref{tbl:bwscheduler}. Here we decrease the bandwidth by 10\% every $L$ epochs. As a result, there is a tradeoff between runtime and performance where the decaying strategy is 1.4-2.6x faster than the original RC-Mixup, but has slightly worse RMSE and MAPE results that are still better than those of the baselines.
\begin{table}[t]
  \caption{\method{} with bandwidth decaying (``With BD'') compared to the \method{} on the real and synthetic datasets.}
  \label{tbl:bwscheduler}
  \centering
  \begin{tabular}{@{\hspace{3pt}}l@{\hspace{4pt}}l@{\hspace{4pt}}c@{\hspace{4pt}}c@{\hspace{4pt}}c@{\hspace{3pt}}}
    \toprule
    Dataset & Method & RMSE & MAPE & Runtime(s) \\
    \midrule
     \multirow{2}{*}{{\sf Spectrum}} &
     \method{}  &  $7.471_{\pm 0.220}$ & $5.930_{\pm 0.165}$ & 578 \\
     &  With BD & $7.903_{\pm 0.144}$ & $6.284_{\pm 0.126}$ & 247 \\
     \midrule
     \multirow{2}{*}{{\sf NO2}} &
     \method{}  & $0.557_{\pm0.034}$ & $13.816_{\pm 1.440}$ & 14  \\
     & With BD & $0.556_{\pm0.029}$ & $13.974_{\pm 1.318}$ & 7 \\
     \midrule
     \multirow{2}{*}{{\sf Airfoil}} &
     \method{}  & $2.530_{\pm 0.357}$ & $1.398_{\pm 0.123}$ & 188 \\
     & With BD & $2.582_{\pm 0.244}$ & $1.446_{\pm 0.111}$ & 73 \\
     \midrule
     \multirow{2}{14mm}{\sf Exchange-Rate} &
     \method{}  & $0.0156_{\pm0.0007}$ & $1.4692_{\pm0.0811}$ & 512 \\
     & With BD & $0.0158_{\pm0.0010}$ & $1.5093_{\pm0.1116}$ & 371 \\
    \bottomrule
  \end{tabular}
\end{table}

\subsection{Varying Noise}

We evaluate how robust \method{} is against different types and levels of noise on the {\sf Spectrum} dataset in Table~\ref{tbl:varyingnoise}. We sample different levels of Gaussian noise by varying the noise magnitude $m$ and also use label flipping. As a result, \method{} consistently performs better than \cmixup{} as it is less affected by noise with its bandwidth tuning. We also evaluate \method{}'s robustness while varying the noise rate on the {\sf Spectrum} dataset from 10\% to 40\% in Table~\ref{tbl:varyingnoise_appendix} and observe results that are consistent with when using the default noise rate of 30\%. 

\begin{table}[t]
\setlength{\tabcolsep}{3pt}
  \caption{\method{} robustness against different types and levels of noise on the {\sf Spectrum} dataset.}
  \label{tbl:varyingnoise}
  \centering
  \begin{tabular}{cclll}
    \toprule
    Noise Type & Magnitude ($m$ value) & Method & RMSE & MAPE\\
    \midrule
    \multirow{7}{*}{Gaussian} & \multirow{2}{*}{Low (1)} & \cmixup{} & 10.044 & 9.031\\
     &  & \method{} & 7.442 & 6.029 \\
     \cmidrule(lr){2-5}
     & \multirow{2}{*}{Medium (2)} & \cmixup{} & 12.125 & 10.840 \\
     &  & \method{} & 7.471 & 5.930 \\
     \cmidrule(lr){2-5}
     & \multirow{2}{*}{High (3)} & \cmixup{} & 13.074 & 11.660 \\
     &  & \method{} & 7.941 & 6.240 \\
     \midrule
     Label & \multirow{2}{*}{n/a} & \cmixup{} & 18.259 & 17.520 \\
     Flipping & & \method{} & 7.893 & 6.124 \\
    \bottomrule
  \end{tabular}
\end{table}

\begin{table}[t]
\setlength{\tabcolsep}{3pt}
  \caption{\method{} robustness against different Gaussian noise rates on the {\sf Spectrum} dataset.}
  \label{tbl:varyingnoise_appendix}
  \centering
  \begin{tabular}{clll}
    \toprule
    Noise Rate & Method & RMSE & MAPE\\
    \midrule
    \multirow{2}{*}{10\%} & \cmixup{} & $9.291_{\pm 0.129}$ & $8.210_{\pm 0.066}$ \\
     & \method{} & $6.100_{\pm 0.118}$ & $5.090_{\pm 0.119}$  \\
     \midrule
     \multirow{2}{*}{20\%} & \cmixup{} & $10.721_{\pm 0.202}$ & $9.583_{\pm 0.232}$ \\
     & \method{} & $6.729_{\pm 0.155}$ & $5.493_{\pm 0.095}$ \\
     \midrule
     \multirow{2}{*}{40\%} & \cmixup{} & $13.376_{\pm 0.369}$ & $12.257_{\pm 0.325}$ \\
     & \method{} & $8.524_{\pm 0.342}$ & $6.784_{\pm 0.293}$ \\
    \bottomrule
  \end{tabular}
\end{table}

\subsection{Parameter Analysis}
\label{sec:parameteranalysis}

\begin{figure}[t]
\centering
    \subfloat[Varying $L$]{
        {\includegraphics[scale=0.23]{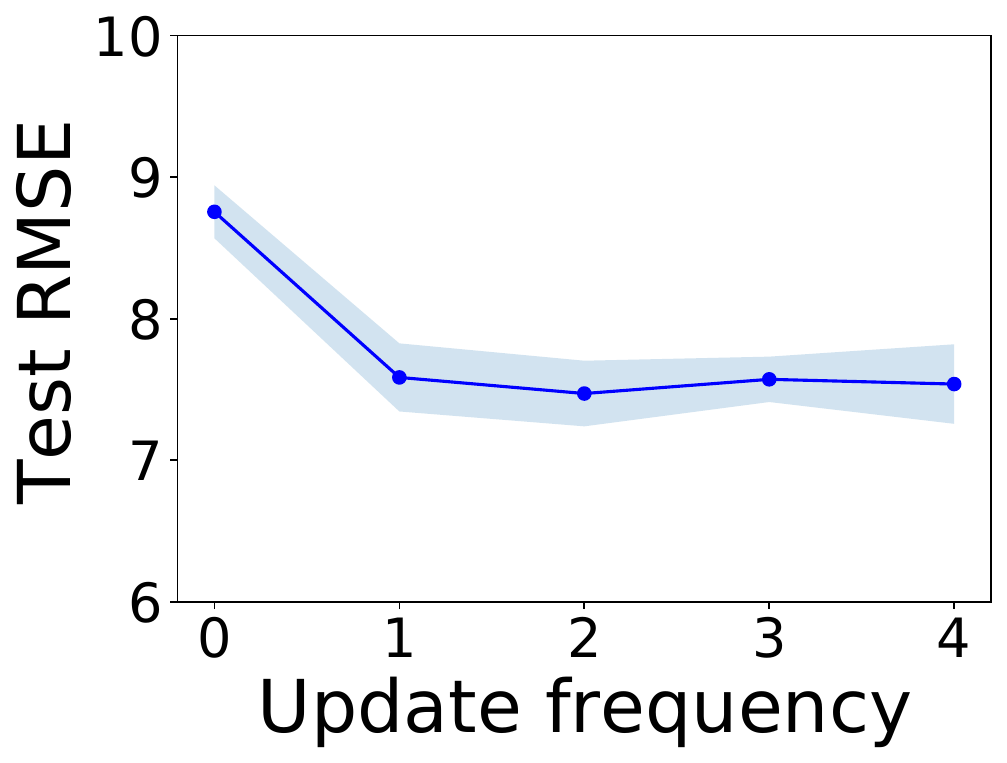}}
        \label{fig:varying_l}
        }
    \subfloat[Varying $N$]{
        {\includegraphics[scale=0.23]{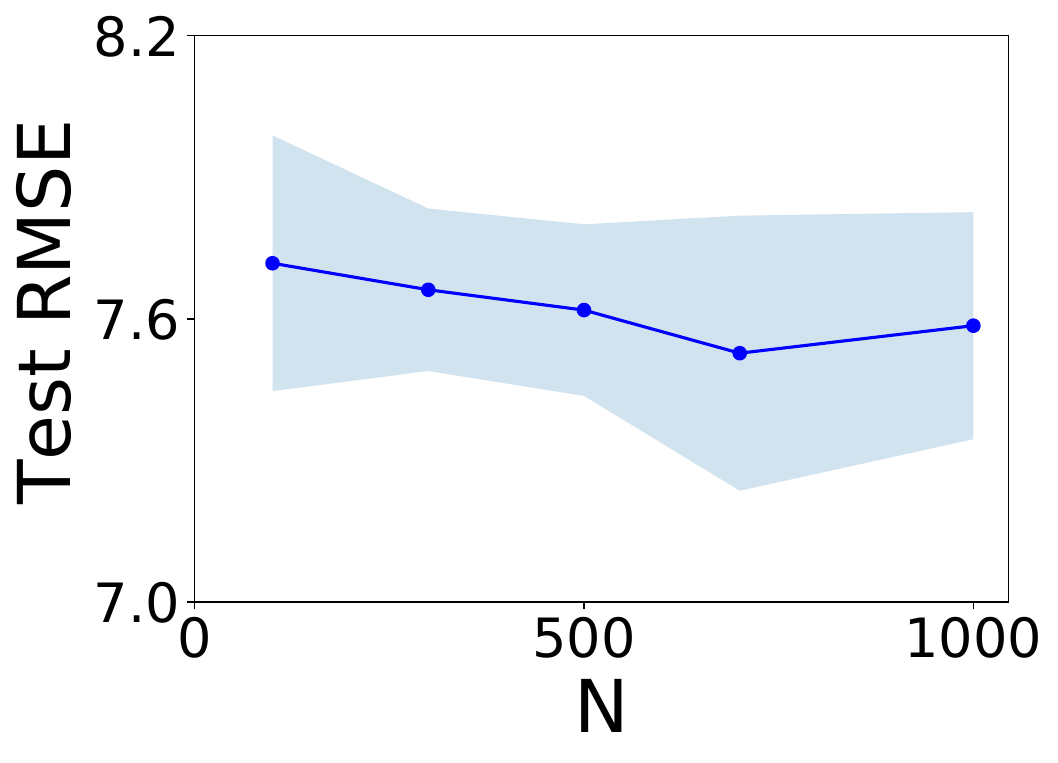}}
        \label{fig:varying_n}
        }
        \vskip -0.05in
\caption{\method{} performance when varying $L$ and $N$ used for bandwidth tuning on the {\sf Spectrum} dataset.}
\label{fig:varyingparams}
\end{figure}

We vary the bandwidth tuning parameters $L$ and $N$ to see how \method{} performs in different scenarios on the {\sf Spectrum} dataset. The results on the other datasets are similar. As we decrease $L$, we update the bandwidth more frequently, which means that it is more likely to be up-to-date. Figure~\ref{fig:varying_l} indeed shows that the RMSE decreases against the number of updates, but only for the first one or two updates. Increasing $N$ means that we evaluate the bandwidths using more rounds before selecting one of them. Figure~\ref{fig:varying_n} shows that a higher $N$ leads to lower RMSE, but with diminishing returns. Hence, \method{} achieves sufficient performance improvements even when both $L$ and $N$ are small, which means that the additional overhead for bandwidth tuning of RC-Mixup is not large.

\subsection{Validation Set Construction and Size}

We evaluate \method{} by varying the validation set size on all the four datasets in Figures~\ref{fig:convergence_bandwidth_validationsetsize}i--\ref{fig:convergence_bandwidth_validationsetsize}l. As a result, \method{} works reasonably well even for smaller validation set sizes.
If there is no clean validation set readily available, we can utilize the given robust training method to construct a validation set ourselves. In Table~\ref{tbl:validationset}, we compare \method{}'s RMSE on two datasets when using (1) a clean validation set; (2) a noisy validation set that is cleaned with robust training first; and (3) a noisy validation set that is a subset of the training set. We compare with (3) just as a reference. As a result, \method{} using (2) has slightly lower, but a comparable performance as when using (1), which means that \method{} can still perform well without a given clean validation set with an additional overhead of cleaning the noisy validation set. %We also show that \method{} performs well against smaller validation set sizes in the appendix.

\begin{table}[t]
  \caption{\method{} performance when using a clean validation set, a noisy validation set that is cleaned with robust training (RT), and a noisy validation set.}
  \label{tbl:validationset}
  \centering
  \begin{tabular}{lccc}
    \toprule
    Dataset & Validation set & RMSE & MAPE \\
    \midrule
     \multirow{3}{*}{{\sf Spectrum}} &
     Clean  &  $7.471_{\pm 0.220}$ & $5.930_{\pm 0.165}$ \\
     & Cleaned with RT & $7.739_{\pm 0.262}$ & $6.209_{\pm 0.225}$  \\
     & Noisy & $8.176_{\pm 0.544}$ & $6.625_{\pm 0.659}$  \\
     \midrule
     \multirow{3}{*}{{\sf Airfoil}} &
     Clean  & $2.530_{\pm 0.357}$ & $1.398_{\pm 0.123}$  \\
     & Cleaned with RT & $2.684_{\pm 0.453}$ & $1.471_{\pm 0.136}$  \\
     & Noisy & $2.692_{\pm 0.244}$ & $1.547_{\pm 0.167}$  \\
    \bottomrule
  \end{tabular}
\end{table}

\subsection{Other Robust Training Methods}
\label{sec:otherrobust}

To show the generality of \method{}, we integrate \cmixup{} with the two other robust training methods O2U-Net\,\cite{DBLP:conf/iccv/HuangQJZ19} and SELFIE\,\cite{DBLP:conf/icml/SongK019} and provide more evaluation results. \method{} can be plugged into any other multi-round robust training method as well.

\paragraph{O2U-Net Integration} O2U-Net consists of two phases for selecting clean samples: a pre-training phase and a cyclical training phase between underfitting and overfitting. During the pre-training phase, we train the model using \cmixup{}. In the cyclical training phase, we further train this pre-trained model by adjusting the learning rate cyclically to obtain the clean samples. We then return the pre-trained \cmixup{} model trained on the final clean samples with bandwidth tuning.

\paragraph{SELFIE Integration} 
SELFIE refurbishes the labels of unclean samples with low predictive uncertainty by assigning them the most frequent class among the previous $q$ predictions. SELFIE was designed for classification, so we extend it to work for regression. 

While SELFIE calculates uncertainty using the entropy of the predictive categorical distribution, this approach cannot be directly applied in a regression setting due to the absence of the categorical distribution. Instead of using entropy, we quantify uncertainty by assessing the variance of predictive labels over the past $q$ predictions. This approach is consistent with other common regression techniques\,\cite{gal2016dropout, lakshminarayanan2017simple}, where the uncertainty corresponds to the variation in predictions across multiple instances or models. 

When refurbishing a label of an unclean sample, we average the previous predictions instead of finding the most frequent label. The reason is that in regression, labels have continuous values, so it makes less sense to find the most frequent labels as in classification.

Finally, since SELFIE refurbishes the labels for every mini-batch, the label distances between two labels change, which means the sampling probabilities for \cmixup{} may change as well. However, updating the distances between label pairs requires significant amounts of computation, and we thus choose not to update the sampling probabilities once they are initially computed. This approach is reasonable because only a fraction of labels that have low uncertainties are actually refurbished. Even if we do update the sampling probabilities, our experiments show that they have a minor impact on the model's performance. Nonetheless, incrementally updating the sampling probabilities efficiently is an interesting future work.

\paragraph{Evaluation} 

We now evaluate \method{} using O2U-Net and SELFIE on the four datasets in Table~\ref{tbl:otherrobusttraining}. As a result, \method{} improves both \cmixup{} and robust training as in Table~\ref{tbl:otherrobusttraining} demonstrating the synergy between the two for all four datasets.

\begin{table}[t]
  \caption{\method{} performance using the robust training methods O2U-Net\,\cite{DBLP:conf/iccv/HuangQJZ19} 
 and SELFIE\,\cite{DBLP:conf/icml/SongK019} on the four datasets.}
  \label{tbl:otherrobusttraining}
  \centering
  \setlength{\tabcolsep}{2pt}
  \begin{tabular}{@{\hspace{2pt}}c@{\hspace{3pt}}lccc@{\hspace{2pt}}}
    \toprule
    Dataset & Method & RMSE & MAPE  \\
    \midrule
    \multirow{6}{*}{{\sf Spectrum}} & \cmixup{} only & $12.125_{\pm 0.200}$ & $10.840_{\pm 0.246}$  \\
    \cmidrule(lr){2-4}
    & O2U-Net only & $9.386_{\pm 0.360}$ & $7.867_{\pm 0.321}$  \\
    & \method{} w/ O2U-Net & $8.372_{\pm 0.316}$ & $6.702_{\pm 0.144}$  \\
    \cmidrule(lr){2-4}
    & SELFIE only & $9.040_{\pm 0.225}$ & $7.000_{\pm 0.178}$  \\
    & \method{} w/ SELFIE & $7.701_{\pm 0.138}$ & $6.234_{\pm 0.186}$  \\
    \midrule
    \multirow{6}{*}{{\sf NO2}} & \cmixup{} only & {$0.586_{\pm 0.033}$} & {$14.604_{\pm 1.283}$}  \\
    \cmidrule(lr){2-4}
    & O2U-Net only & {$0.564_{\pm 0.033}$} & {$14.052_{\pm 1.617}$}  \\
    & \method{} w/ O2U-Net & {$0.554_{\pm 0.022}$} & {$13.782_{\pm 1.285}$}  \\
    \cmidrule(lr){2-4}
    & SELFIE only & {$0.559_{\pm 0.043}$} & {$14.002_{\pm 1.580}$}  \\
    & \method{} w/ SELFIE & {$0.550_{\pm 0.030}$} & {$13.562_{\pm 1.247}$}  \\
    \midrule
    \multirow{6}{*}{{\sf Airfoil}} & \cmixup{} only & {$3.438_{\pm0.218}$} & {$2.093_{\pm 0.185}$}  \\
    \cmidrule(lr){2-4}
    & O2U-Net only & {$3.026_{\pm0.199}$} & {$1.783_{\pm 0.073}$}  \\
    & \method{} w/ O2U-Net & {$2.795_{\pm0.379}$} & {$1.574_{\pm 0.145}$}  \\
    \cmidrule(lr){2-4}
    & SELFIE only & {$2.713_{\pm0.426}$} & {$1.508_{\pm 0.174}$}  \\
    & \method{} w/ SELFIE & {$2.569_{\pm0.358}$} & {$1.408_{\pm 0.165}$}  \\
    \midrule
    \multirow{6}{16mm}{\sf Exchange-Rate} & \cmixup{} only & {$0.0216_{\pm 0.0018}$} & {$2.2931_{\pm0.2153}$}  \\
    \cmidrule(lr){2-4}
    & O2U-Net only & {$0.0203_{\pm 0.0015}$} & {$2.0909_{\pm0.1676}$}  \\
    & \method{} w/ O2U-Net & {$0.0162_{\pm 0.0011}$} & {$1.5523_{\pm0.1407}$}  \\
    \cmidrule(lr){2-4}
    & SELFIE only & {$0.0165_{\pm 0.0004}$} & {$1.5741_{\pm0.0666}$}  \\
    & \method{} w/ SELFIE & {$0.0154_{\pm 0.0009}$} & {$1.4535_{\pm0.1137}$}  \\
    \bottomrule
  \end{tabular}
\end{table}

\section{Related Work}
\label{sec:relatedwork}

There are largely two branches of work for data augmentation in regression. One is semi-supervised regression\,\citep{DBLP:journals/jifs/KostopoulosKKR18,zhou2005semi,DBLP:journals/eswa/KangKC16,kim2020reliability} where the goal is to utilize unlabeled data for training. Another branch is data augmentation when there is no unlabeled data, which is our research focus. Most data augmentation techniques are tailored to classification and more recently a few are designed for regression.

\paragraph{Data Augmentation for Classification} There are largely three approaches: generative models, policies, and Mixup techniques. Generative models including GANs\,\citep{DBLP:conf/nips/GoodfellowPMXWOCB14} and VAEs\,\citep{DBLP:journals/corr/KingmaW13} are popular in classification where the idea is to generate realistic data that cannot be distinguished from the real data by a discriminator. Another approach is to use policies\,\citep{DBLP:conf/cvpr/CubukZMVL19}, which specify fixing rules for transforming the data. However, a major assumption is that the labels of these generated samples are the same, which does not necessarily hold in a regression setting where most samples may have different labels. Mixup\,\citep{DBLP:conf/iclr/ZhangCDL18, DBLP:conf/iccv/YunHCOYC19, kim2020puzzle, kim2021comixup, verma2019manifold} takes the alternative approach of generating both data and labels together by mixing existing samples with different labels assuming linearity between training samples\,\citep{DBLP:conf/nips/ChapelleWBV00,onthewu2020}.

\paragraph{Data Augmentation for Regression} 

%Categorize so we can see multi-round robust training as a category
There is a new and increasing literature on data augmentation for regression using Mixup techniques. Although the original Mixup paper mentions that its techniques can easily be extended to regression, the linearity of a regression model is limited where mixing samples with all others may not be beneficial and even detrimental to model performance. Hence, the key is to figure out how to limit the mixing. 
RegMix\,\cite{hwang2022regmix} learns for each sample how many nearest neighbors in terms of data distance it should be mixed with for the best model performance using a validation set. 
The state-of-the-art \cmixup{}\,\cite{DBLP:conf/nips/YaoWZZF22} uses a Gaussian kernel to generate a sampling probability distribution for each sample based on label distances using a global bandwidth and selects a sample to mix based on the distribution. Anchor Data Augmentation\,\cite{schneider2023anchor} takes a more domain-specific approach where it clusters data points and modifies original points either towards or away from the cluster centroids for the augmentation. Finally, R-Mixup\,\cite{kanrmixup2023} specializes in improving model performance on biological networks. In comparison, \method{} uses the domain-agnostic \cmixup{} and makes it robust against noise.

\paragraph{Robust Training} 

The goal of robust training is to train accurate models against noisy or adversarial data, and we cover the representative works. While data can be problematic in various places\,\cite{xiao2015learning}, most techniques assume that the labels are noisy\,\cite{song2020learning} and mitigate with the following strategies: (1) developing model architectures that are less affected by the noise\,\cite{chen2015webly, jindal2016learning, bekker2016training, han2018masking}, (2) applying loss regularization techniques to reduce overfitting\,\cite{43405, pereyra2017regularizing, tanno2019learning, hendrycks2019using, Menon2020Can}, (3) correcting the loss function to account for the noise\,\cite{patrini2017making, NIPS2017_2f37d101, ma2018dimensionality, arazo2019unsupervised}, and (4) proposing sample selection techniques\,\cite{DBLP:conf/icml/ShenS19,DBLP:conf/icml/SongK019,DBLP:conf/iccv/HuangQJZ19} for selecting clean samples from noisy data. In comparison, \method{} is mainly designed to work with the sample selection techniques, but can also complement other approaches as long as it can access intermediate clean data within rounds.

\section{Conclusion}
We proposed \method{}, an effective and practical data augmentation strategy against noisy data for regression. The state-of-the-art data augmentation technique \cmixup{} is not designed to handle noise, and \method{} is the first to tightly integrates it with robust training for a synergistic effect. \cmixup{} benefits from the intermediate clean data information identified by robust training and performs different mixing depending on whether noisy data is being used, while robust training cleans its data better with \cmixup{}'s data augmentation. We also proposed dynamic tuning techniques for \cmixup{}'s bandwidth. Our extensive experiments showed how \method{} significantly outperforms \cmixup{} and robust training baselines on noisy data benchmarks and is compatible with various robust training methods. 

\section*{Acknowledgement}
This work was supported by Samsung Electronics Co., Ltd., by the Institute of Information \& Communications Technology Planning \& Evaluation(IITP) grant funded by the Korea government(MSIT) (No.\@ 2022-0-00157, Robust, Fair, Extensible Data-Centric Continual Learning), and by the National Research Foundation of Korea(NRF) grant funded by the Korea government(MSIT) (No.\@ NRF-2022R1A2C2004382)
% This work is supported by the Samsung Electronics' University R\&D program [Algorithm advancement and reliability-enhancing technique development for spectrum data-based deep learning].

\bibliographystyle{ACM-Reference-Format}
\balance
\bibliography{main}

\clearpage

\onecolumn
\appendix

\section{Appendix -- Experiments}

\subsection{Other Experimental Settings}

We explain more detailed experimental setups.

\paragraph{Hyperparameters} 
We summarize the hyperparameters for all the datasets in Table~\ref{tbl:hyperparameters}. For each dataset, we either use the traditional notion of Mixup or the more recent Manifold Mixup\,\citep{verma2019manifold} (ManiMix), whichever performs better. We use a 3-layer fully connected neural network with one hidden layer with 128 nodes (FCN3) for the {\sf Spectrum}, {\sf NO2}, and {\sf Airfoil} datasets. For the {\sf Exchange-Rate} dataset, we build an LST-Attn model\,\citep{lai2018modeling} with a horizon value of 12.

\begin{table}[h]
\setlength{\tabcolsep}{2.5pt}
  \caption{Detailed hyperparameters for each dataset used in the experiments.}
  \label{tbl:hyperparameters}
  \centering
  \begin{tabular}{lcccc}
    \toprule
    Dataset & {\sf Spectrum} & {\sf NO2} & {\sf Airfoil} & {\sf Exchange-Rate} \\
    \midrule
     Mixup type & Mixup & Mixup & ManiMix & Mixup \\
     Batch size & 128 & 32 & 16 & 128 \\
     Learning rate & 1e-2 & 1e-2 & 1e-2 & 1e-3 \\
     Maximum epochs & 2,000 & 200 & 400 & 200 \\
     $B$ & \{5, 10, 15, 20\} & \{1e-3, 5e-3, 1e-2, 5e-2, 1e-1, 5e-1, 1\} & \{1e-3, 1e-2, 1e-1, 1, 10\} & \{1e-3, 1e-2, 5e-2, 1e-1\} \\
     Default bandwidth & 20 & 2 & 10 & 2 \\
     $L$ & 500 & 100 & 200 & 50 \\
     $N$ & 500 & 100 & 200 & 50 \\
     $\alpha$ for Mixup & 2.0 & 2.0 & 0.5 & 2.0 \\
     Noise magnitude $m$ & 2 & 4 & 2 & 5 \\
     Model architecture & FCN3 & FCN3 & FCN3 & LST-Attn \\
     Optimizer & Adam & Adam & Adam & Adam \\
     
    \bottomrule
  \end{tabular}
\end{table}

\subsection{Noisy Versus Hard-to-train Data}
While noisy data tends to have high loss values, so does hard-to-train data. However, noisy data typically has higher losses as shown in Paul et al.\,\citep{paul2021deep}. Here hard data is identified using the EL2N score, which is similar to loss. Paul et al.\,\citep{paul2021deep} analyze the challenge of distinguishing hard data from noisy data and shows that data with noisy labels tends to have higher EL2N scores. If there is a label noise, data with large loss values are commonly considered as noise \,\citep{DBLP:conf/icml/ShenS19, DBLP:conf/iccv/HuangQJZ19, DBLP:conf/icml/SongK019}.

\end{document}